\documentclass[journal,12pt,onecolumn,draftclsnofoot]{IEEEtran}

\usepackage[margin=1in]{geometry}
\usepackage{amsmath,amssymb,mathtools}
\usepackage{booktabs,array}
\usepackage[colorlinks=true,linkcolor=blue,citecolor=blue,urlcolor=blue]{hyperref}
\usepackage{comment}
\usepackage{mathrsfs}
\usepackage{graphicx}
\usepackage{amsmath}        
\usepackage{amssymb}        
\usepackage{amsthm}
\usepackage[shortlabels]{enumitem}
\usepackage{graphicx}
\theoremstyle{plain}
\newtheorem{theorem}{Theorem}[section] 
\newtheorem{mtheorem}{Meta Theorem}

\newtheorem{mcorollary}{Meta Corollary}
\newtheorem{proposition}[theorem]{Proposition}

\theoremstyle{definition}
\newtheorem{definition}[theorem]{Definition}

\theoremstyle{remark}
\newtheorem*{remark}{Remark}  
\newtheorem{assumption}{Assumption}

\newcommand{\inputlen}{{t_0}}
\newcommand{\ahead}{m}

\newcommand{\E}{\mathbb{E}}
\newcommand{\CE}{\mathbb{E}}

\newcommand{\KL}{D_{\mathrm{KL}}}

\newcommand{\Law}{\mathcal{L}}
\newcommand{\del}{\partial}
\newcommand{\Xt}{X^{\theta}}
\newcommand{\phisol}{\phi_z^\theta(t)}
\newcommand{\CLd}{L_{data}}
\newcommand{\CO}{\mathcal{O}}
\newcommand{\reals}{\mathbb{R}}
\newcommand{\bdim}{d_b}
\newcommand{\xdim}{d_x} 
\newcommand{\tf}{f}
\newcommand{\wct}{\tau}
\newcommand{\wcn}{N_c}
\newcommand{\V}{}

\newcommand{\N}{\mathbb{N}}

\newcommand{\CX}{\mathcal{X}}
\newcommand{\CL}{\mathcal{L}}
\newcommand{\Xd}{X_{data}}
\newcommand{\lip}{b}
\newcommand{\CA}{\mathcal{A}}
\newcommand{\ps}{p^*}
\newcommand{\phisoln}{\phi_z^{\theta}} %
\newcommand{\vvar}{v} %
\newcommand{\bo}{\boldsymbol{1}} %
\newcommand{\nat}{\mathbb{N}} %
\newcommand{\om}{\omega} %
\newcommand{\CW}{\Delta}
\newcommand{\belief}{\pi}
\newcommand{\emb}{{W}_{\text{emb}}}
\newcommand{\softmax}{\operatorname{softmax}}
\newcommand{\cn}{q} 
\newcommand{\AR}{f}
\newcommand{\Embed}{E}
\newcommand{\I}{\mathcal{C}}
\newcommand{\defn}{\stackrel{\text{defn}}{=}}
\newcommand{\unit}{\mathbf{1}}
\newcommand{\Belief}{\Pi(V)}
\newcommand{\Simplex}{\Pi}
\newcommand{\FFN}{\operatorname{FFN}}
\newcommand{\Layer}{\mathsf{Layer}}
\newcommand{\LayerNorm}{\operatorname{LayerNorm}}
\newcommand{\Attn}{\operatorname{Attn}}
\newcommand{\MultiHead}{\operatorname{MultiHead}}
\newcommand{\nhead}{n_{\text{head}}}

\newenvironment{myassumptions}{%
   \begin{description}[style=multiline, leftmargin = 18pt, align=left,font=\normalfont]%
}{%
   \end{description}%
}
 \makeatletter
  \def\nl#1#2{\begingroup
     \scalebox{0.85}[1]{\textbf{#2}}%
    \def\@currentlabel{\textnormal{\scalebox{0.85}[1]{\textbf{#2}}}}
     \phantomsection\label{#1}\endgroup
}
\makeatother

\title{Stochastic Dynamics of Generative AI: From Autoregressive Models to Interacting LLM Networks and Neural SDEs}
\author{Vikram Krishnamurthy, Yiming Zhang,  Luke Snow\thanks{The authors contributed equally to the paper. Vikram Krishnamurthy vikramk@cornell.edu  and Yiming Zhang yz2926@cornell.edu are with the School of Electrical and Computer Engineering, Cornell University, Ithaca,  NY 14853, USA. Luke Snow luke.snow@yale.edu is with the Department of Electrical Engineering, Yale University, New Haven, CT 06520, USA. \\ This research was supported by
NSF grant CCF-2312198 and Army Research office grant W911NF-24-1-0083.}}
\date{}

\begin{document}

\maketitle

\begin{abstract}
This paper develops a stochastic dynamical-systems
perspective on three classes of generative AI models.
First, we formulate large language models (LLMs) as
high-dimensional nonlinear autoregressive models
with attention-based dependencies. This formulation
connects pretraining by conditional maximum likelihood
to sequential token generation and provides a 
probabilistic interpretation of LLM operation.
Second, we consider networks of interacting LLMs
and characterize their collective behavior through
mean-field ordinary differential equations.
These dynamics provide a framework for understanding
emergent preferential attachment and persistent
structural biases, including type-dependent
glass-ceiling effects.
Third, we consider neural stochastic differential
equations (SDEs) as continuous-time generative models for
 stochastic trajectories, including those  observed at irregular time intervals. We discuss  their training
 with
particular emphasis on Wiener-space cubature methods
based on rough-path analysis as an efficient alternative to Monte Carlo estimation.
Together, the three settings address  probabilistic prediction, information exchange and emergent behavior in interacting  networks, and  path-space generative modeling.
\end{abstract}

\section{Introduction}
Generative AI models learn probabilistic representations
of data and use them to generate new samples, such as
text, images, and stochastic trajectories.
A broad class of generative AI
models can be viewed as stochastic dynamical systems
whose parameters are learned from data. From this
perspective, generation is a sequential probabilistic
process in which the current state and its history
determine the distribution of future states.
This perspective connects modern generative AI to probabilistic prediction, information exchange in interacting networks, and the analysis and numerical approximation of stochastic processes.

This tutorial paper presents  a unified dynamical-systems
perspective on generative AI through three
complementary settings:
(i) large language models (LLMs) as high-dimensional
nonlinear autoregressive models;
(ii) networks of interacting LLMs and their
mean-field dynamics; and
(iii) neural stochastic differential equations
(SDEs) as continuous-time generative models.
Although these settings differ in their state spaces,
interaction mechanisms, and time scales, they share
a common mathematical foundation: stochastic
evolution governed by learned nonlinear functions.
Our treatment is selective:
we emphasize
modeling and 
underlying stochastic mechanisms
 rather than
a broad survey of the literature.
Accordingly,
we cite only selected references.


{\bf Part 1}.
\textbf{LLMs as high-dimensional nonlinear autoregressive (AR) models.}
LLMs based on transformer architectures
generate text by sequentially sampling tokens from conditional
probability distributions determined by the preceding token history.
Part 1 discusses the formulation of LLMs as
high-dimensional, high-order nonlinear autoregressive (AR)
models with attention-based dependencies.
Here, ``dimension'' refers to the dimension of the
latent hidden state produced by the transformer
(typically $10^3$--$10^4$), while ``order'' refers
to the number of past tokens available for predicting
the next token, as determined by the context window
(which can be of order $10^6$).
We describe how the nonlinear AR model is constructed
through causal self-attention and how its conditional
probability distributions are estimated by maximum-likelihood
pretraining and subsequently used for token generation.
The exposition emphasizes mathematical abstraction
over architectural detail, with brief remarks on
alignment and inference.


\textbf{Part 2. Mean-Field Dynamics and Structural Bias in Networks of LLMs.}
Motivated by sociological phenomena in human social
networks, Part~2 models interacting LLMs as networks
of coupled nonlinear AR systems.
We characterize their collective
evolution through a mean-field ordinary differential
equation (ODE) obtained in the large-network limit,
under suitable assumptions.

The mean-field dynamics provide a framework for
analyzing emergent network structure and structural
bias. We study how preferential
attachment and type-dependent interactions  lead
to disparities in connectivity, including
glass-ceiling effects. The equilibria and stability
properties of the mean-field ODE characterize the
long-run behavior of the interacting agents.

\textbf{Part 3. Neural SDEs as continuous-time generative models.}
While Parts 1 and 2  focus on LLMs and networks of LLMs (discrete-time
autoregressive models and their interactions), many other applications
of generative AI involve continuous-time stochastic processes
observed at irregular time intervals.
Neural stochastic differential equations (SDEs) provide a
flexible generative framework for such time series by
parameterizing the drift and diffusion coefficients with
neural networks.
These models generate probability distributions on path space,
with applications including financial time-series simulation,
modeling human motion trajectories, and reconstructing
gene-regulatory dynamics from noisy, irregularly sampled data.

In Part 3 we introduce  neural SDEs. After motivating their usefulness, we introduce the powerful method of Wiener space cubature and illustrate its use in  training neural SDEs. 
Wiener space cubature replaces the Brownian motion driving an SDE with a finite collection of judiciously chosen deterministic trajectories, yielding an ordinary differential equation for each trajectory.
Drawing on rough-path analysis, these trajectories and their associated weights are constructed to match moments of iterated Brownian integrals, so that a weighted convex combination of the resulting solutions approximates the desired expectation.
They provide improvements over the default Monte Carlo estimator, with the capability of root mean-squared error of $O(n^{-1})$ instead of $O(n^{-1/2})$. Neural SDEs can be considered as a form of generative AI for continuous-time stochastic processes; they  generalize \emph{finite-dimensional} generative models, such as score-based diffusion, to a \emph{function space}.

The three parts of this paper 
 connect probabilistic prediction,
information exchange in interacting networks,
and efficient computation of expectations under
continuous-time generative models. 
They illustrate how nonlinear  AR  models
describe stochastic token generation in LLMs,
how interactions among LLMs produce emergent
network structure and bias, and how Wiener-space
cubature provides an alternative to Monte Carlo
estimation in neural SDE training.


\section{Part 1. LLMs as high-dimensional nonlinear autoregressive  models}

This section formulates LLMs
as high-dimensional nonlinear autoregressive (AR) models
with attention-based dependencies, and explains how
their conditional probability distributions are learned
through pretraining.

\paragraph*{How does an  LLM generate tokens?}
As a short appetizer, we first outline how
an LLM generates new tokens. An LLM can be viewed as a $\CW$-order
nonlinear AR model, where $\CW$  denotes   \index{context window}
the context window length. After training, the optimized parameters $\theta^\star$ of the transformer neural  network are fixed.
The user-provided \textit{prompt},  consists of the initial tokens
$c=x_{1:t_0}=\{x_1,\ldots,x_{\inputlen}\}$. For $t \ge \inputlen$, subsequent tokens
$x_{t+1}, x_{t+2}, \ldots$ are generated autoregressively by the LLM as follows:
At time $t$, the LLM uses the previous up to $\CW$ tokens\footnote{Indices corresponding to $t<1$ are omitted.}
$x_{t-\CW+1:t}$ to predict the next token. These tokens constitute the
model's \textit{context}.  \index{context}
The LLM then draws the next token $x_{t+1}$ randomly from the
$V$-dimensional probability vector $\belief_t^\tau$
(where $V$ denotes the size of the LLM vocabulary):
\begin{align}\label{eq:AR_gen}
\begin{split}
&x_{t+1}
\sim
\belief_t^\tau
=
\softmax\!\left(
\frac{W^*_{\mathrm{out}}\,
      \AR_{\theta^\star,t}\!\bigl(x_{t-\CW+1:t}\bigr)
      + b^*_{\mathrm{out}}}{\tau}
\right).
\end{split}
\end{align}
 In the above AR model, 
$\tau> 0$  is the user-chosen sampling temperature, and  $W^*_{\mathrm{out}}\in \mathbb{R}^{V \times d}$, $b^*_{\mathrm{out}}\in \mathbb{R}^V$ are learned  parameters, where $d$ denotes the latent hidden-state (embedding) dimension of the transformer. Also for any $z\in \reals^V$, $\softmax(z) $ is the $V$-dimensional probability vector with elements $\exp(z_i)/\sum_j \exp(z_j)$, $i=1\ldots, V$.

The AR feature map $\AR_{\theta,t}\in \reals^d$ appearing in~\eqref{eq:AR_gen} is  the final-layer hidden state corresponding to the most
recent token $t$, defined by the composition of
 $L$ layers of the transformer neural network:
\begin{align}
\begin{split}\label{eq:AR_comp}
&\AR_{\theta,t}\!\bigl(x_{t-\CW+1:t}\bigr)
\;\triangleq\;
h_{L,t},
\quad
h_{\ell,t}  = g^{(\ell)}_\theta\!\bigl(h_{\ell-1,t-\CW+1}, \ldots  ,h_{\ell-1,t} \bigr), \\&
\quad \ell=1,\ldots,L,
\end{split}
\end{align}
initialized at layer 0 by
$$ (h_{0,t-\CW+1}, \ldots, h_{0,t})  = \emb[x_{t-\CW+1:t}]$$
where $\emb \in \reals^{d \times V}$ denotes the learned token embedding
matrix that maps each discrete token to $\reals^d$. 
We refer to  $h_{\ell,s} \in \reals^d$ as the hidden-state representation outputted at layer $\ell$ for token~$s$.  Detailed constructions are given in~\eqref{eq:hidden_state_matrices}, \eqref{eq:attention_scores}.

An important 
feature of~\eqref{eq:AR_comp} is  the 
self-attention mechanism in each layer~$\ell$: 
\begin{equation}\label{eq:g_layer}
g^{(\ell)}_\theta\!\bigl(h_{\ell-1,t-\CW+1},\ldots,h_{\ell-1,t}\bigr)
=
\phi\!\left(
\sum_{s=t-\CW+1}^{t}
\softmax_s\!\bigl(a^{(\ell)}_\theta(t)\bigr)\,
L^{(\ell)}_\theta\, h_{\ell-1,s}
\right),
\end{equation}
where $a^{(\ell)}_\theta(t)\in \reals^\CW$ denotes the vector
of attention scores with real-valued elements 
\[ a^{(\ell)}_\theta(t,r)
\;\defn\;
\big\langle M^{(\ell)}_\theta\, h_{\ell-1,t},\;
          N^{(\ell)}_\theta\, h_{\ell-1,r}\big\rangle,
\qquad r=t-\CW+1,\ldots,t, \]
and  $\softmax_s(z)$ denotes the $s$-th component of the $\CW$-dimensional 
probability vector $\softmax(z)$.
Here,  $M^{(\ell)}_\theta$, $N^{(\ell)}_\theta$ and $L^{(\ell)}_\theta$ are
learned matrices.  In~\eqref{eq:g_layer}, $\phi(\cdot)$ denotes a nonlinear transformation
of the attention output. It 
is a schematic representation of a transformer layer;
the full architecture also includes residual connections, normalization, and feedforward operations,
see~\eqref{eq:hidden_state_matrices}. 


{\bf Context}. 
There are two main takeaways: First, tokens are drawn from probability vector~\eqref{eq:AR_gen} specified by a $\CW$-order AR model, where the dimension of the hidden state is $d$. Second,  inside the softmax operator lies the 
all-important  attention mechanism~\eqref{eq:g_layer} which is a bilinear function of the hidden
states $h$ inside the softmax, followed by a linear  (weighted-sum) aggregation   of~$h$.
Remarkably,  this  bilinear–softmax–linear mechanism, when composed repeatedly
across  multiple layers of a transformer neural network, makes an LLM  appear almost  sentient, despite
the absence of explicit symbolic reasoning or  predesigned task-specific user interface.

By the chain rule of probability, the distribution
of a token sequence factors into conditional
next-token distributions. The causal transformer
parametrizes these distributions through a
high-dimensional nonlinear AR map with attention.
Pretraining estimates the parameters of this
conditional probability kernel by maximizing the
likelihood of observed token sequences. Once
trained, the same kernel generates new sequences
by recursively sampling  the next token and
updating the context. 

Training an LLM  has  two stages:
\begin{itemize}
  \item {\em 1. Pretraining},  where the LLM learns its core language ability by 
    predicting the next token on a large text corpus;
\item  \emph{2. Alignment training}, 
where the pretrained LLM is fine-tuned using human feedback to make its 
behavior helpful, honest, and safe.
\end{itemize}

Pretraining and alignment constitute the main
stages of LLM training. Pretraining learns the
conditional next-token distributions from a large
corpus, while alignment subsequently modifies these
distributions to favor responses satisfying human
preferences or other objectives. Once trained,
the LLM generates text by recursively sampling
from its learned conditional distributions.
We will focus  on the  formulation of
pretraining, with brief remarks on alignment
and generation.

\subsection{Vocabulary, Corpus and Token Embeddings} 

   \paragraph{Vocabulary}
The LLM has fixed vocabulary
\[
\mathcal{V} = \{ w_1, w_2, \dots, w_V \},
\]
where the vocabulary size  $V$ is typically around $100,000$  Byte Pair Encoding (BPE) tokens.  \index{tokens!BPE} \index{BPE (Byte Pair Encoding)}
A BPE token is generally smaller than  a word and larger than a character.

\paragraph{Training Corpus}
The training corpus for an LLM comprises $N$ documents (sequences). Each document is a  collection of token sequences where each token\footnote{We identify each token in the vocabulary $ \mathcal{V} = \{ w_1, w_2, \dots, w_V \}$ with its integer index in $\{1,\ldots, V\}$. Thus, $x_t^{(n)}$ denotes the index of the token at position $t$ in sequence $n$.}
  $x^{(n)}_t \in \mathcal{V}$.
\[
\mathcal{D}
=
\Big\{
(x^{(n)}_1, \dots, x^{(n)}_{T_n})
\;\big|\;
n=1,\dots,N
\Big\},
\]
with $T_n$ the length of document $n$.  
LLMs train on  total number of tokens $\sum_{n=1}^N T_n$ of $1$–$20$ trillion tokens.

\paragraph{Token embeddings.}
The learned embedding matrix $\emb\in\reals^{d\times V}$ maps a token index into the continuous feature vector
\begin{equation}\label{eq:token_embedding}
e_t^{(n)}=\emb[x_t^{(n)}]\in\reals^d, \quad t= 1,\ldots, T_n.
\end{equation}
For sequence $n$, comprising  $T_n$ tokens, stack the embeddings as columns into the matrix
\begin{equation}\label{eq:embedding_matrix}
E^{(n)}=[e_1^{(n)},\ldots,e_{T_n}^{(n)}]
=
\emb[x_{1:T_n}^{(n)}]
\in\reals^{d\times {T_n}}.
\end{equation}
These vectors are the input to the causal transformer; their dimension $d$ is distinct from the vocabulary size $V$ and the context length $\CW$.

\subsection{Causal Transformer Architecture as a Nonlinear Map}

\paragraph{Hidden Latent State.}
A causal transformer with parameters $\theta$ is a
 nonlinear map
\[
F_\theta:\reals^{d\times T_n}
\longrightarrow\reals^{d\times T_n},
\]
composed of masked self-attention and feedforward layers.
For a sequence of length $T_n\leq\CW$, it produces
the hidden-state matrix
\begin{equation}\label{eq:transformer_1}
S^{(n)}
=
F_\theta(\Embed^{(n)})
=
[s_1^{(n)},\ldots,s_{T_n}^{(n)}]
\in\reals^{d\times T_n}.
\end{equation}
Each column $s_t^{(n)}$ is a context-dependent
representation of token $t$. The transformer
processes the sequence in parallel, but
\emph{causal masking} ensures that $s_t^{(n)}$
depends only on tokens $x_{1:t}^{(n)}$.

Let $\unit_t$ denote the $t$-th unit vector
in $\reals^{T_n}$. Then
\begin{equation}\label{eq:causal_masking}
s_t^{(n)}
=
F_\theta(\Embed^{(n)})\,\unit_t,
\qquad
\frac{\partial s_t^{(n)}}
{\partial (e_r^{(n)})^\top}
=0_{d\times d},\quad r>t.
\end{equation}
Thus, causal masking imposes a
\emph{block lower-triangular structure} on
the Jacobian of $F_\theta$, with each block
of dimension $d\times d$.

For an arbitrary position $t$ in a longer sequence,
recall the context window
$\I_t=\{\max(1,t-\CW+1),\ldots,t\}$.
Evaluating the same causal transformer on this
truncated context gives the nonlinear AR representation
\begin{equation}\label{eq:causal_masking2}
s_t^{(n)} 
=
\AR_{\theta,t}(x_{\I_t}^{(n)})
\defn
F_\theta\bigl(\emb[x_{\I_t}^{(n)}]\bigr)
\,\unit_{|\I_t|}
\in\reals^d.
\end{equation}
Here $F_\theta$ is applied to an input of
length $|\I_t|$, with the corresponding
positional information. By construction,
\[
\frac{\partial s_t^{(n)}}
{\partial (e_r^{(n)})^\top}
=0_{d\times d},
\qquad r\notin\I_t.
\]
Hence the induced mapping from a full token
sequence to its hidden states has a
\emph{banded block lower-triangular Jacobian},
with bandwidth $\CW$ in token positions.
The context restriction applies to the entire
transformer computation, across all its layers.

\paragraph{Output Layer: Logits.}
The hidden state is mapped to a vector of
next-token logits:
\begin{equation}\label{eq:next_logit}
z_t^{(n)}
=
W_{\mathrm{out}}s_t^{(n)}+b_{\mathrm{out}}
\in\reals^V,
\end{equation}
where $W_{\mathrm{out}}\in\reals^{V\times d}$
and $b_{\mathrm{out}}\in\reals^V$ are learned
parameters.

The explicit construction of $F_\theta$,
including multi-head causal self-attention
and Rotary Position Embeddings (RoPE),
is developed below.

\subsection{Pretraining: Conditional Maximum Likelihood}
\label{subsec:pretraining}

The standard cross-entropy
pretraining objective is  the negative
log-likelihood associated with the nonlinear
AR model introduced above.

\paragraph{Softmax Prediction.}
Given the logit vector $z_t^{(n)}\in\reals^V$
in~\eqref{eq:next_logit}, the predicted next-token
probability distribution is
\begin{equation}\label{eq:softmax_predict}
\hat{\belief}_t^{(n)}
\defn
\softmax(z_t^{(n)})\in\Belief,
\qquad
\hat{\belief}_t^{(n)}(w_i)
=
\frac{\exp(z_{t,i}^{(n)})}
{\sum_{j=1}^V\exp(z_{t,j}^{(n)})},
\end{equation}
where $\Belief$ denotes the probability simplex
over the $V$ tokens. Thus,
$\hat{\belief}_t^{(n)}(w_i)$ is the conditional
probability assigned to token $w_i$ given the
observed context $x_{\I_t}^{(n)}$.

\paragraph{Conditional Maximum Likelihood.}
The AR formulation provides a direct
connection between pretraining and generation.
By the chain rule for probabilities, the conditional
likelihood of an observed token sequence is
\begin{equation}\label{eq:ar_likelihood}
p_\theta(x_{2:T_n}^{(n)}\mid x_1^{(n)})
=
\prod_{t=1}^{T_n-1}
\hat{\belief}_t^{(n)}
\bigl(x_{t+1}^{(n)}\bigr).
\end{equation}
Here each conditional probability depends on at
most the preceding $\CW$ tokens.

Pretraining estimates the model parameters by
minimizing the negative log-likelihood over the
training corpus:
\begin{equation}\label{eq:loss_entropy}
\mathcal L(\theta)
=
-\sum_{n=1}^N\sum_{t=1}^{T_n-1}
\log\hat{\belief}_t^{(n)}
\bigl(x_{t+1}^{(n)}\bigr)\\
=
\sum_{n=1}^N\sum_{t=1}^{T_n-1}
\KL\!\left(
\unit_{x_{t+1}^{(n)}},
\hat{\belief}_t^{(n)}
\right),
\end{equation}
where $\unit_{x_{t+1}^{(n)}}$ denotes the
$V$-dimensional indicator vector corresponding
to the observed next token.
$\KL(\alpha,\beta)$ denotes the
Kullback--Leibler divergence between probability
vectors $\alpha$ and~$\beta$.

The maximum likelihood estimate of the model parameters is  obtained by 
solving
\[
\min_\theta\mathcal L(\theta),
\qquad
\theta=
\bigl(
\emb,\,
\text{transformer-layer weights},\,
W_{\mathrm{out}},\,
b_{\mathrm{out}}
\bigr).
\]

In pretraining, the conditional distributions
are evaluated using the token histories
from the training corpus, rather than tokens
generated by the model. Causal masking allows
these conditional distributions to be evaluated
in parallel across token positions in each
training sequence.

The key observation is that pretraining estimates
the conditional probability kernel of the nonlinear
AR model, whereas generation recursively samples
from the learned kernel, as in~\eqref{eq:AR_gen}.
Thus, pretraining and generation are two
complementary uses of the same stochastic AR model.

\subsubsection*{Summary (Nonlinear AR Model)}

For each token position $t$ in document $n$, the LLM predicts the distribution of the next
token $x^{(n)}_{t+1}$ as  \index{nonlinear AR model}
\[
\hat{\belief}^{(n)}_t
=
\softmax\!\left(
W_{\text{out}}\, s^{(n)}_t + b_{\text{out}}
\right) \in \Belief, 
\qquad
s^{(n)}_t =  F_\theta(\Embed^{(n)}) \,\unit_t = F_\theta(\emb[x_{1:t}^{(n)}]) \,\unit_t \in \reals^{d}.
\]
Notice that  prediction  of $x^{(n)}_{t+1}$  uses the entire prefix $x^{(n)}_{1:t}$. Hence the above representation is a large
($\CW$-order)  nonlinear   AR  model, where $\CW$ is the context window length. We  write it more transparently as 
$$ x^{(n)}_{t+1} \sim \hat{\belief}^{(n)}_t = \softmax\bigg(
W_{\text{out}}\, \AR_{\theta,t}\bigl(x^{(n)}_{t-\CW+1:t}\bigr) + b_{\text{out}}
\bigg),
$$
where $\AR_{\theta,t}$  defined in~\eqref{eq:causal_masking2} is a deterministic nonlinear state construction function  induced by the transformer. 
The structure of $\AR_{\theta,t}$ in terms of self attention is elaborated  below.

\subsection{Multi-Head Self-Attention} \label{app:rope}

In~\eqref{eq:transformer_1}, the transformer neural network  was defined  abstractly as a nonlinear
map
\[
F_\theta:\;\Embed^{(n)} \longmapsto S^{(n)}\in\reals^{d\times T_n},
\qquad
s_t = S^{(n)}\unit_t,
\]
where the hidden state matrix  $s_t^{(n)}$ outputted by the transformer was 
used to produce the conditional distribution of the next token using~\eqref{eq:next_logit} and~\eqref{eq:softmax_predict}. 
We now specify the internal structure of $F_\theta$.

\paragraph{Layered Structure of $F_\theta$.}
The transformer $F_\theta$ is the composition of $L$ layers,
\[
F_\theta = \Layer^{(L)}\circ\cdots\circ\Layer^{(1)},
\qquad
H_\ell=\Layer^{(\ell)}(H_{\ell-1}),\quad H_0=\Embed^{(n)}.
\]
Here, $H_\ell\in \reals^{d\times T_n}$ denotes the  hidden-state matrix \textit{after} layer $\ell$,
 with
\[
H_\ell=[h_{\ell,1},\ldots,h_{\ell,T_n}],
\quad
s_t^{(n)}=h_{L,t},
\quad S^{(n)}=H_L.
\]
Thus $h_{\ell,t}$ is the intermediate hidden state
at layer $\ell$ and token position $t$, while
$s_t^{(n)}$ is the final-layer hidden state.
We suppress the document index $n$ on $h_{\ell,t}$
for notational simplicity.
Note that $H_0 = \Embed^{(n)}$ are  the input token embeddings.

Each layer $\ell$ consists of a self-attention block followed by a position-wise
feed-forward network, denoted  $\FFN_\ell$, applied to each token's hidden state:
\begin{equation}\label{eq:hidden_state_matrices}
\begin{aligned}
\tilde H_\ell
&=
H_{\ell-1}
+
\MultiHead_\ell\!\bigl(\LayerNorm_\ell(H_{\ell-1})\bigr)  \in \reals^{d\times T_n}, 
\\
H_\ell
&=
\tilde H_\ell
+
\FFN_\ell\!\bigl(\LayerNorm'_\ell(\tilde H_\ell)\bigr),
\qquad \ell=1,\dots,L.
\end{aligned}
\end{equation}
Thus $F_\theta(\Embed^{(n)})=H_L$, namely the output of the $L$-th (final) layer of the transformer.

Here $\LayerNorm_\ell$ denotes layer normalization: for $h\in\reals^d$, it is defined by
\[
\LayerNorm_\ell(h)
=
\gamma_\ell\odot
\frac{h-\bar h\mathbf{1}}
{\sqrt{d^{-1}\|h-\bar h\mathbf{1}\|^2+\epsilon}}
+\beta_\ell,
\qquad
\bar h=\frac{1}{d}\mathbf{1}^\top h,
\]
where $\odot$ denotes element-wise multiplication, $\gamma_\ell,\beta_\ell\in\reals^d$ are
learned parameters and $\epsilon>0$ ensures
numerical stability. The primed operator
$\LayerNorm'_\ell$ has the same form with
separate independent learned parameters.

\paragraph{Multi-Head Self-Attention with Bilinear Scores}  \index{self-attention}

 Self-attention lets the current token look back at all previous tokens within the context window and automatically learn --  from their content --  how much attention to pay to each one when predicting the next token.
Multi-head attention runs many such mechanisms in parallel and combines them.

Each layer $\ell$ has $\nhead$ attention heads. 
Fix a layer $\ell$, attention head $\ahead \in \{1,\ldots,\nhead\}$, and let $d_h=d/\nhead$.
For each token $t$  and context window length $\CW$, attention is computed over the \textit{context window} $\I_t$ of previous tokens: \index{context window}
\[\I_t \defn \{\max\{1,t-\CW+1\},\ldots,t\}, \text{ where } |\I_t| = \Delta .
\]
%
Then for each token $t$ and any earlier token $s \in I_t$, define the query, key and value vectors as 
\[
q_{\ell,t}^\ahead = W_Q^{\ahead,\ell} h_{\ell-1,t} \in \reals^{d_h},\quad
k_{\ell,s}^\ahead = W_K^{\ahead,\ell} h_{\ell-1,s} \in\reals^{d_h},\quad
v_{\ell,s}^\ahead = W_V^{\ahead,\ell} h_{\ell-1,s} \in\reals^{d_h}.
\]
The query vector $q_{\ell,t}^\ahead$, key vector $k_{\ell,s}^\ahead$,  value vector  $v_{\ell,s}^\ahead$, and state $ h_{\ell-1,t}$ depend on the training document  $n$; we suppress this index to simplify notation.
In contrast,
$W_Q^{\ahead,\ell},W_K^{\ahead,\ell},W_V^{\ahead,\ell}\in\reals^{d_h\times d}$ are the learned 
query, key and value projection matrices, respectively, shared across all training documents (and not $n$ dependent).

The (causal) attention scores are defined as \emph{bilinear functions of the hidden
states} $(h_{\ell-1,t},h_{\ell-1,s})$:  \index{self-attention!bilinear form}
\begin{equation}\label{eq:attention_scores}
a_{\theta,\ell}^\ahead(t,s)
=
\frac{(q_{\ell,t}^\ahead)^\top (R_t^\top R_s)\,k_{\ell,s}^\ahead}{\sqrt{d_h}} = \frac{
h_{\ell-1,t}^\top
\Big(
(W_Q^{\ahead,\ell})^\top R_t^\top R_s\, W_K^{\ahead,\ell}
\Big)
h_{\ell-1,s}
}{\sqrt{d_h}},
\qquad s \in \I_t,
\end{equation}
where the matrices $R_u \in \reals^{d_h \times d_h}$ implement rotary position embeddings (defined below).  The attention scores are unconstrained real-valued scalars; nonnegativity and normalization arise after applying the softmax, yielding the attention weights that form a $\CW$-dimensional probability vector:
\begin{equation}
  \label{eq:causal_attention}
w_{\theta,\ell}^\ahead(t,\I_t)=\softmax\!\bigl(a_{\theta,\ell}^\ahead(t,\I_t)\bigr)\in\Simplex(\CW).
\end{equation}
Finally  the $\ahead$-th attention  head output at token $t$ in layer $\ell$ is 
\begin{equation}
  \label{eq:boxedattn}
\Attn_\ahead^\ell(h_{\ell-1,t})
=
\sum_{s \in \I_t} w_{\theta,\ell}^\ahead(t,s)\,v_{\ell,s}^\ahead
=
\sum_{s \in \I_t}
\softmax_s\!\left( 
\frac{ \big[ R_t\, W_Q^{\ahead,\ell} h_{\ell-1,t}\big]^{\top}\,
 R_s
W_K^{\ahead,\ell} h_{\ell-1,s}
}{\sqrt{d_h}}
\right)
\;
\bigl(W_V^{\ahead,\ell} h_{\ell-1,s}\bigr).
\end{equation}
Note that $\Attn_\ahead^\ell:
\reals^d \to \reals^{d_h}$ where  $d_h={d}/{\nhead}$.
For token 
$t$, and any vector $z(t,\I_t)  \in \reals^{\CW}$,  $\softmax_s(z(t,s))$ is the $s$-th component of
 probability vector $\softmax(z(t,\I_t) )\in \Simplex(\CW)$.

Concatenating all $\nhead$ heads and applying an output projection matrix $W_O^\ell$ gives
\[
\MultiHead_\ell(X)
=
W_O^\ell
\begin{bmatrix}
\Attn_1^\ell(X)\\
\vdots\\
\Attn_{\nhead}^\ell(X)
\end{bmatrix}
\in\reals^{d\times T_n},
\qquad W_O^\ell\in\reals^{d\times d}.
\]
The above attention scores, weights, and multi-head attention
outputs  depend on the training document $n$
through its hidden states. We suppressed the 
index $n$  for notational simplicity.

{\em Remarks:} 1.  The hidden state 
$h$ enters the attention mechanism in~\eqref{eq:boxedattn}  three times: as query, key, and value.
Thus, each attention head implements a \emph{bilinear similarity kernel}
followed by a \emph{linear state aggregation}.

2. The attention score $a_{\theta,\ell}^\ahead(t,s)$ is best viewed as a learned similarity score: how much does the current token $t$ match the earlier token $s$. The larger the score, the more token $s$ will contribute to predicting the next token. 

3. {\em  Matrix Form of Attention (Per-Head  Decomposition).}
Multi-head attention is obtained by applying the following construction
for each head $\ahead\in \{1,\ldots,\nhead\}$ and then concatenating the
results followed by the output projection $W_O^\ell$:
Stack the rotated query and  key vectors, and the  value vectors,
over all $T_n$ tokens of the $n$-th training document:
\[
Q_\ahead \;\defn\; 
\begin{bmatrix}
R_1 q_{\ell,1}^\ahead & \cdots & R_{T_n} q_{\ell,T_n}^\ahead
\end{bmatrix}
\in\reals^{d_h\times T_n},
\qquad
K_\ahead \;\defn\;
\begin{bmatrix}
R_1 k_{\ell,1}^\ahead & \cdots & R_{T_n} k_{\ell,T_n}^\ahead
\end{bmatrix}
\in\reals^{d_h\times T_n},
\]
\[
V_\ahead \;\defn\;
\begin{bmatrix}
v_{\ell,1}^\ahead & \cdots & v_{\ell,T_n}^\ahead
\end{bmatrix}
\in\reals^{d_h\times T_n},
\qquad d_h \defn d/\nhead .
\]
Then the attention output for head $\ahead$ over the entire sequence can be written
compactly as: \index{self-attention!matrix form}
\[
 \text{Given } H_{\ell-1} \in \reals^{d \times T_n}, \quad 
\Attn_\ahead^\ell(H_{\ell-1})
=
V_\ahead\,
\softmax\!\Biggl(
\frac{K_\ahead^\top Q_\ahead}{\sqrt{d_h}}
\Biggr)
\;\in\;\reals^{d_h\times T_n}.
\]
The softmax is applied column-wise. In column $t$,
the entries indexed by $s\in\I_t$ are the
attention weights $w_{\theta,\ell}^\ahead(t,s)$;
all other entries are zero.

Causality and the context-window restriction are
enforced by setting the entries $(s,t)$ of
$K_\ahead^\top Q_\ahead$ to $-\infty$ whenever
$s>t$ or $s<t-\CW+1$, before applying
the column-wise $\softmax$.


4. {\em Rotary Position Embeddings (RoPE)}: Because transformers contain no recurrence or convolution, self-attention is  \index{Rotary Position Embeddings (RoPE)} \index{transformer architecture!RoPE}  \index{positional encoding!RoPE}
permutation-equivariant with respect to token order unless positional information is injected; positional
encodings allow the model to represent word order.

Rotary Position Embeddings (RoPE) incorporate positional
information into the attention scores.
In~\eqref{eq:attention_scores},
$R_t\in\reals^{d_h\times d_h}$ denotes an orthogonal
rotation matrix associated with token position~$t$.
These matrices satisfy
$
R_t^\top R_s=R_{s-t},
$,
so that the bilinear attention scores depend on the
relative positions $s-t$ of tokens, without adding
positional vectors to the input embeddings.

\subsection{Alignment training and Generation}
\paragraph{Alignment training}
Pretraining estimates the parameters $\theta$ of the
conditional next-token distribution
$p_\theta(x_{t+1}\mid x_{1:t})$ by maximizing the
likelihood of the training corpus. Alignment
subsequently modifies these parameters to favor
responses satisfying additional criteria, such as
human preferences or verifiable rewards.
Methods include reinforcement learning from human
feedback (RLHF), direct preference optimization
(DPO), and reinforcement learning from verifiable
rewards (RLVR). Although their objectives differ,
these methods preserve the underlying AR
formulation: alignment changes the conditional
probability kernel rather than the mechanism of
sequential token generation. See
\cite{krishnamurthy2026llms} for details.

\paragraph{Generation}
After training and alignment, the parameters are
fixed 
at $\theta^\star$, 
and the LLM generates a response conditioned
on the initial prompt $c=x_{1:\inputlen}$ by successively sampling
from the AR kernel.
The probability distribution over the
generated tokens $x_{\inputlen+1:t}$ is
\begin{equation}
p_{\theta^\star}(x_{\inputlen+1:t}\mid c)
=
\prod_{j=\inputlen+1}^{t}
p_{\theta^\star}(x_j\mid x_{1:j-1}),
\qquad t>\inputlen.
\label{eq:ar_path_distribution}
\end{equation}
Sampling temperature and other decoding rules
modify the generation procedure without changing
the learned parameters. Similarly, in-context
learning and chain-of-thought prompting modify
the conditioning history rather than  model
parameters. This stochastic dynamical viewpoint
extends  to networks of interacting LLMs,
where the generated tokens of one agent enter the
conditioning context of other agents, as
discussed in Part~2.

\begin{figure*}[t!]
\centering
\includegraphics[width=0.8\textwidth]{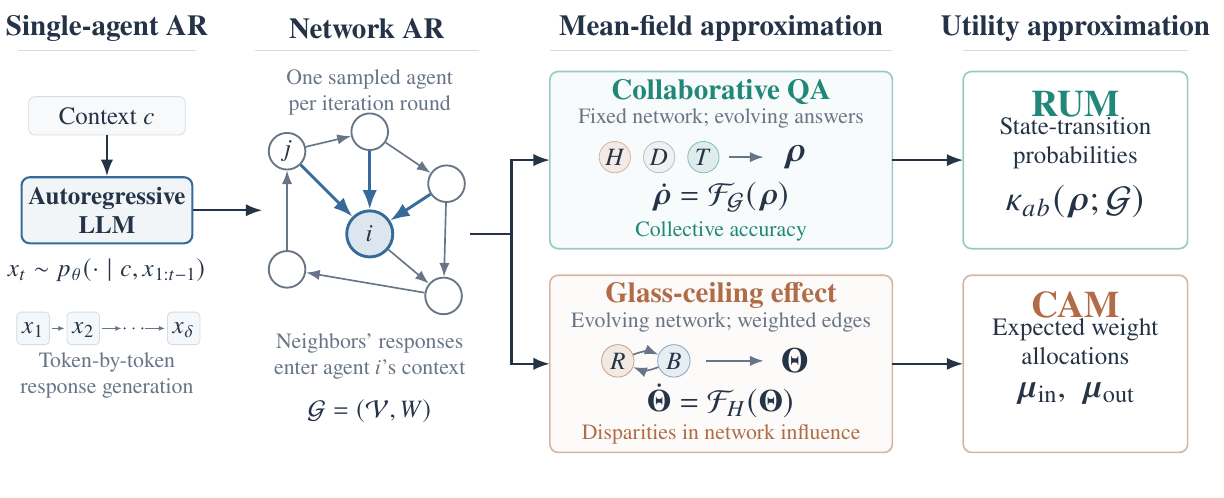}
\vspace{-5mm}
    \caption{
    Overview of the modeling framework:
    from single-agent autoregressive (AR) generation
    to collective dynamics in LLM networks.
    Interactions couple agents' AR processes through
    their contexts, while mean-field approximations
    describe population-level dynamics.
    For collaborative Q\&A, the random utility model (RUM) approximates the
    transition probabilities $\kappa_{ab}$ governing
    response-state proportions.
    For evolving networks, the cross-attention model (CAM) approximates the expected
    weight allocations $\boldsymbol\mu_{\mathrm{in}}$
    and $\boldsymbol\mu_{\mathrm{out}}$ governing
    type-level centrality and glass-ceiling effects.
    }
\vspace{-3mm}
\label{fig:section_teaser}
\end{figure*}

\section{Part 2. Mean-Field Dynamics and Structural Bias in Networks of LLMs}
\label{sec:llm_networks}

Part 1 (previous section) formulated LLMs as
high-dimensional nonlinear autoregressive (AR)
models with attention-based dependencies.
In this section, we extend the formulation
to networks of interacting LLMs, modeled as
coupled nonlinear AR models.
Fig.~\ref{fig:section_teaser} summarizes the
modeling framework, connecting individual
token generation and network interactions
to mean-field dynamics and utility-based
approximations.
The key results of this section are:
\begin{enumerate}
    \item A mean-field ordinary differential
    equation (ODE) model describing the evolution
    of opinions among interacting LLM agents.
    \item A characterization of structural bias
    in interacting LLM networks through the
    stable fixed points of the mean-field ODE,
    revealing conditions under which a
    glass-ceiling effect emerges.
\end{enumerate}


Following the AR interpretation in
Part 1, an LLM processes and
generates text as sequences of discrete tokens
from a fixed vocabulary.
Let $c=(c_1,\ldots,c_\inputlen)$ denote the input prompt (context),
a sequence of $m$ tokens encoding the instructions,
query, and any additional information provided
to the model.
Let $x_t$ denote the single token generated at
output position $t$, and let
$x_{1:t-1}=(x_1,\ldots,x_{t-1})$ denote the
sequence of previously generated tokens.\footnote{In Part~1, the prompt consisted of the tokens
$x_1,\ldots,x_{t_0}$, and the generated response
begins with $x_{t_0+1}$. In Part~2, for notational convenience we denote
the prompt separately by $c=(c_1,\ldots,c_{t_0})$
and restart the token indexing for the generated
response at $x_1$.}
Conditioned on $c$ and $x_{1:t-1}$, the LLM
generates the next token according to
\begin{equation}
    x_t
    \sim p_\theta(\cdot\mid c,x_{1:t-1})
    =
    \operatorname{softmax}\left(
        \frac{
            W_{\mathrm{out}}
            f_{\theta,t-1}(c,x_{1:t-1})
            +b_{\mathrm{out}}
        }{\tau}
    \right).
    \label{eq:overview_token_ar}
\end{equation}

Fig.~\ref{fig:section_teaser} summarizes the  modeling framework of Part~2.
Starting from a single-agent autoregressive (AR) model, we
incorporate interactions among agents to obtain a network AR model.
A mean-field approximation then describes the evolution of
aggregate system states. Finally, RUM or CAM approximates the
transition probabilities or expected increments governing these
dynamics, providing a unified framework for collaborative Q\&A
and the glass-ceiling effect.

\subsection{Network of LLM Agents as Interacting AR Model}

In this section, we extend the AR model of
a single LLM to a network of interacting LLM agents.
By an \textit{LLM agent}, we mean a prompted LLM
of a specified base model, e.g., GPT-6 or Gemini-4.
Such agents increasingly operate within larger
systems, where they interact by exchanging
messages or consuming content generated by
one another \cite{jain2025interacting}.
Privacy requirements, finite context windows,
and computational costs can limit each agent's
access to information, motivating networks
in which agents combine partial observations
through communication.
Applications include conversational workflows
\cite{wu2023autogen}, collaborative reasoning
and debate \cite{du2024improving}, and distributed
processing of long documents \cite{zhang2024chain}.
As one agent's response enters another's context,
their AR generation processes become coupled,
enabling information sharing but also potentially
reinforcing errors and producing herding
\cite{jain2025interacting}.
The network determines who influences whom,
while agents' choices of interaction partners
can in turn reshape the network
\cite{papachristou2025network}.
An interacting AR model captures these dependencies
by specifying how neighboring responses enter
each agent's context and influence its next output.
It thus provides a common framework for studying
information propagation over fixed networks
and the feedback between agent-generated decisions
and network evolution.

Therefore, we extend the above AR model to a network
of interacting LLM agents.
Consider a directed network $\mathcal G=(\mathcal V,W)$,
where $\mathcal V=\{1,\ldots,N\}$ is the set of agents
and $W$ is the set of directed edges.
An edge $(j,i)\in W$ means that agent $i$ can receive
information from agent $j$, and
$\mathcal N_i=\{j\in\mathcal V:(j,i)\in W\}$
denotes its incoming neighbors.
Agents interact through random asynchronous updates,
following standard models of distributed interaction
and opinion dynamics
\cite{elboim2024asynchronous}.
At each interaction round, one agent is selected
independently and uniformly from $\mathcal V$.
The selected agent generates a new response using
the most recent responses of its incoming neighbors,
while the responses of all other agents remain unchanged.
These neighboring responses may have been generated
at different earlier events, so agents do not need
to update in synchronized rounds.
This schedule has an equivalent continuous-time
interpretation: each agent has an independent
Poisson clock with the same rate and updates whenever
its clock rings.
Observing the process at successive clock events
therefore yields independent, uniformly distributed
agent selections.

Let $x_t^i(k+1)$ denote the $t$-th token generated by
agent $i$ at interaction round $k+1$, and let
$x^i(k)$ denote the token sequence representing
agent $i$'s current response after interaction round  $k$,
with token indices suppressed.
At each interaction round $k+1$, one agent $i$
is sampled to receive the current responses of
all its incoming neighbors and generate its next
response token by token:
\begin{equation}
    x_t^{i}(k+1)
    \sim
    p_{\theta_i}\left(
        \cdot\mid c^i(k),x_{1:t-1}^{i}(k+1)
    \right),
    \qquad k\geq0.
    \label{eq:overview_network_ar}
\end{equation}
Here, the pmf $p_{\theta_i}$ is defined by
\eqref{eq:overview_token_ar} with the transformer
parameters $\theta_i$ of agent $i$.
All other agents retain their current responses.
The initial prompt $c^i(0)$ differs across agents
and remains fixed throughout the interaction process.
For interaction round $k\geq1$, the context used by the sampled
agent $i$ is
\begin{equation}
    c^i(k)
    =
    \{
        c^i(0),x^i(k),
        \{x^j(k):j\in\mathcal N_i\}
    \},
    \qquad k\geq1.
    \label{eq:overview_network_context}
\end{equation}
Here, $\mathcal N_i=\{j\in\mathcal V:(j,i)\in W\}$
denotes the set of incoming neighbors of agent $i$.
The braces specify the information included in
the context: the initial context $c^i(0)$,
agent $i$'s current response $x^i(k)$,
and its neighbors' current responses $x^j(k)$.
These inputs are formatted into an ordered token
sequence and remain fixed during generation,
while the generated prefix $x_{1:t-1}^i(k+1)$
grows as new tokens are generated.
Thus, each agent follows a token-level AR
model whose context incorporates responses from
its neighbors, coupling the agents' generation
processes across rounds.

\subsection{Example 1. Collaborative Q\&A}

In this section, we study collaborative question-and-answering (collaborative Q\&A) through information propagation in a network of interacting LLM agents.

\subsubsection{Why Collaborative Q\&A?}
Collaborative Q\&A has become increasingly important
for long-context reasoning \cite{levy2024same}, addressing concerns
about sensitive-data leakage \cite{song2024securesql}, and enabling
collaboration among LLM agents with different roles \cite{estornell2024multi}.
Collaborative Q\&A 
encourages distributing
document processing across LLM agents that exchange
intermediate responses, as demonstrated by
multi-agent approaches to long-context reasoning
\cite{zhang2024chain}.
However, communication does not necessarily
improve accuracy: repeated interactions can
reinforce shared misconceptions and lead agents
toward an incorrect consensus
\cite{estornell2024multi}.
The main question is: 
\begin{quote}
How does  network
structure, agent capability, and the distribution
of private evidence jointly shape the propagation
of information and hallucinations?
\end{quote}

We consider LLM agents answering a shared question
$q$, with each agent $i$ receiving private evidence
$E_i$.
Both $q$ and $E_i$ are token sequences encoding
the question and the evidence available to agent $i$,
respectively.
For example, $q$ encodes ``Where is Alice's book?'',
$E_1$ encodes ``Alice put her book in the drawer,''
and $E_2$ encodes ``Alice left the room.''
Here, $E_1$ provides sufficient information to
answer $q$, whereas $E_2$ alone does not.
Agent $i$'s initial context is
$c^i(0)=\{q,E_i\}$, formatted into an ordered
token sequence.

\subsubsection{Motivation for Mean Field Dynamics}

Mean-field dynamics describe the collective behavior of
a large population of interacting stochastic agents
through a deterministic equation governing the evolution
of their empirical state distribution in the
large-population limit.

To address the above  question, we construct a mean-field dynamic ordinary differential equation (ODE)  to  model  the evolving
proportions of correct, uncertain, and hallucinating LLM agents. The main outcome is two fold: First, by estimating the parameters of the ODE from real LLM agents data, we can use the ODE as a parsimonious generative model  to predict the behavior of the network as it evolves over time. Second, this mean field ODE has a unique stable fixed point (under reasonable conditions)  -- this facilitates predicting the asymptotic fraction of nodes that are truthful, uncertain or hallucinating for any initial condition.

\subsubsection{From Interacting AR model to Mean Field Dynamics}

The agents interact over a fixed network
$\mathcal G$ with fixed neighbor sets $\mathcal N_i$.
Specializing \eqref{eq:overview_network_ar} to this
setting, the agent $i$ selected at interaction round $k+1$
generates its next token sequence according to
\begin{equation}
\begin{gathered}
    x_t^i(k+1)
    \sim p_{\theta_i}\left(
        \cdot\mid c^i(k),x_{1:t-1}^i(k+1)
    \right),
    \qquad k\geq0,\\[4pt]
    c^i(k)=
    \left\{
        q,E_i,x^i(k),
        \{x^j(k):j\in\mathcal N_i\}
    \right\},
    \qquad k\geq1.
\end{gathered}
\label{eq:qa_network_ar}
\end{equation}

Each agent $i$ generates its initial response from
$c^i(0)$ and subsequently incorporates its own
current response and those of its neighbors.
The question and private evidence remain fixed
throughout these interactions.
Let $\mathcal X$ denote the set of finite token
sequences over the model's vocabulary.
We classify each response token sequence using
\begin{equation}
    \psi:\mathcal X\to\{H,D,T\},
    \qquad
    \psi(x^i(k))
    =
    \begin{cases}
        H, & \text{hallucinated response},\\
        D, & \text{uncertain response (``don't know'')},\\
        T, & \text{correct response}.
    \end{cases}
    \label{eq:qa_response_state}
\end{equation}
Hallucination refers broadly to generated content
that is factually incorrect or unfaithful to
the source information \cite{ji2023survey}.
In our three-state model, we operationalize $H$
as a response that provides an incorrect answer,
assessed against the question's ground truth.
For the example above, ``The book is on the table''
is classified as $H$, ``I don't know'' as $D$,
and ``The book is in the drawer'' as $T$.
This classification uses ground truth for analysis,
which is not supplied to the agents.

We define the expected fraction of LLM agents
in each response state as
\begin{equation}
    \rho_b^k
    =
    \frac{1}{|\mathcal V|}
    \sum_{i\in\mathcal V}
    \Pr\!\left(\psi(x^i(k))=b\right),
    \qquad b\in\{H,D,T\}.
    \label{eq:qa_expected_fractions}
\end{equation}
Here, $|\mathcal V|$ denotes the total number
of agents in the network.
We collect these fractions into the population vector
\begin{equation}
    \boldsymbol\rho^k
    =
    (\rho_H^k,\rho_D^k,\rho_T^k)^\top,
    \qquad
    \rho_H^k+\rho_D^k+\rho_T^k=1.
    \label{eq:qa_population_vector}
\end{equation}
Its components represent the expected fractions
of hallucinated, uncertain, and correct responses,
respectively.
The population state space is
\[
    \mathcal S
    =
    \left\{
        \boldsymbol\rho\in\mathbb R_+^3:
        \rho_H+\rho_D+\rho_T=1
    \right\}.
\]
Let $\widehat{\boldsymbol\rho}^k$ denote the
empirical population fractions, with
$\widehat\rho_b^k
=|\mathcal V|^{-1}\sum_{i\in\mathcal V}
\mathbf 1\{\psi(x^i(k))=b\}$,
so that $\boldsymbol\rho^k
=\mathbb E[\widehat{\boldsymbol\rho}^k]$.
Under the mean-field closure, their expectations
evolve according to
\begin{equation}
    \boldsymbol\rho^{k+1}
    =
    \boldsymbol\rho^k
    +
    \frac{1}{|\mathcal V|}
    \mathbb E\!\left[
        \mathcal F_{\mathcal G}
        \bigl(\widehat{\boldsymbol\rho}^k\bigr)
    \right].
    \label{eq:qa_expected_recursion}
\end{equation}
The factor $1/|\mathcal V|$ arises because only
one agent updates at each interaction round.
We rescale interaction time as
\begin{equation}
    s=\frac{k}{|\mathcal V|},
    \qquad k\geq1.
    \label{eq:qa_time_scaling}
\end{equation}
Under suitable regularity and initial-condition
assumptions (equicontinuity and boundedness of the
interpolated sequences so that the Arzel\`a--Ascoli
theorem applies, together with Lipschitz continuity
of $\mathcal F_{\mathcal G}$), the deterministic
large-population limit satisfies the ODE
\begin{equation}
    \frac{d\boldsymbol\rho(s)}{ds}
    =
    \mathcal F_{\mathcal G}
    \bigl(\boldsymbol\rho(s)\bigr),
    \qquad s\in\mathbb R_+.
    \label{eq:qa_mean_field}
\end{equation}
Here,
$\boldsymbol\rho(s)=(\rho_H(s),\rho_D(s),\rho_T(s))^\top$
is the population state as
$|\mathcal V|\to\infty$.

\subsubsection{Random Utility Model (RUM)}

Our next task is to obtain a tractable empirical
approximation to the mean-field dynamics
$\mathcal F_{\mathcal G}$. For this purpose,
we introduce the random utility model (RUM), widely used in
microeconomics. RUM
provides a
parametric framework (typically logistic model) for estimating utility
functions from observed stochastic choices
of agents in a population.
RUM models 
discrete
choice by assigning each alternative a deterministic
utility plus a random perturbation and selecting
the alternative with the highest realized utility
\cite{mcfadden1974conditional,train2009discrete}.
Here, the alternatives are the response states
$H$, $D$, and $T$.
For a sampled agent currently in state $a$,
the deterministic utilities depend on the population
state and network structure, while the perturbations
capture variability in response outcomes.
Independent standard Gumbel perturbations yield
multinomial logit transition probabilities,
whose utility parameters can be estimated from
observed transitions by maximum likelihood
\cite{train2009discrete}.

We use RUM to approximate the state-transition
probabilities induced by LLM interactions,
thereby obtaining an approximation to the drift
$\mathcal F_{\mathcal G}$.
We detail this approximation and the mean-field
derivation in Appendix~\ref{subsec:collaborative_qa}
and use the resulting dynamics to study the group's
equilibrium accuracy.
For example, the $H$-component of the drift is
the net rate at which agents enter state $H$.
Let $\kappa_{ab}(\boldsymbol\rho;\mathcal G)$ denote,
under the mean-field closure, the probability
that an activated agent currently in state $a$
transitions to state $b$, given the population
state $\boldsymbol\rho$.
Then,
\begin{equation}
    \mathcal F_{\mathcal G,H}(\boldsymbol\rho)
    =
    \rho_D \kappa_{DH}(\boldsymbol\rho;\mathcal G)
    + \rho_T \kappa_{TH}(\boldsymbol\rho;\mathcal G)
    - \rho_H\bigl[
        \kappa_{HD}(\boldsymbol\rho;\mathcal G)
        + \kappa_{HT}(\boldsymbol\rho;\mathcal G)
    \bigr].
\end{equation}
The first two terms describe transitions into
state $H$, while the last term describes
transitions out of state $H$.
RUM is used to approximate the transition
probabilities $\kappa_{ab}$, thereby yielding
an approximation to the mean-field drift.
The drift components for states $D$ and $T$
are obtained in the same way.


\subsubsection{Stability of Mean Field Dynamics}
We next state sufficient conditions for the
closed RUM population model to converge to its
mean-field ODE and for that ODE to approach
a unique equilibrium.
Let
$\kappa_{ab}^{\mathrm{RUM}}
(\boldsymbol\rho;\mathcal G)$
denote the RUM approximation of the averaged
state-transition probability, with
$\sum_b\kappa_{ab}^{\mathrm{RUM}}
(\boldsymbol\rho;\mathcal G)=1$.
The corresponding drift is
\[
    \mathcal F_{\mathcal G,b}^{\mathrm{RUM}}
    (\boldsymbol\rho)
    =
    \sum_{a\in\{H,D,T\}}
    \rho_a
    \kappa_{ab}^{\mathrm{RUM}}
    (\boldsymbol\rho;\mathcal G)
    -\rho_b,
    \qquad b\in\{H,D,T\}.
\]
To make the population-size dependence explicit,
we write $\widehat{\boldsymbol\rho}_N^k$
for the empirical population vector in a model
with $N=|\mathcal V|$ agents.
For the large-population limit, we consider
closed population models governed by the same
averaged transition kernel.

The RUM approximation admits a conditional
universal approximation justification.
For the sampled agent $i$, the AR model
\eqref{eq:qa_network_ar} induces the response-state
probabilities
\[
    \Pr\!\left(
        \psi(x^i(k+1))=b\mid c^i(k)
    \right)
    =
    \sum_{\substack{x\in\mathcal X\\\psi(x)=b}}
    \prod_{t=1}^{|x|}
    p_{\theta_i}\!\left(
        x_t\mid c^i(k),x_{1:t-1}
    \right),
\]
where the sum runs over completed responses,
including the end-of-sequence token.
Thus, each state probability is obtained by
summing the AR probabilities of all responses
classified into that state.
Under the mean-field closure, averaging over
sampled agents and their contexts, conditional
on the current state $a$ and population state
$\boldsymbol\rho$, yields
$\kappa_{ab}(\boldsymbol\rho;\mathcal G)$.

For strictly positive transition probabilities,
a RUM with independent standard Gumbel
perturbations and deterministic utilities
$u_{ab}(\boldsymbol\rho;\mathcal G)
=\log\kappa_{ab}(\boldsymbol\rho;\mathcal G)$
reproduces these probabilities exactly:
\[
    \frac{
        \exp(u_{ab}(\boldsymbol\rho;\mathcal G))
    }{
        \sum_{z\in\{H,D,T\}}
        \exp(u_{az}(\boldsymbol\rho;\mathcal G))
    }
    =
    \kappa_{ab}(\boldsymbol\rho;\mathcal G).
\]
Subtracting $u_{aD}$ from all utilities preserves
these probabilities and sets the reference
utility to zero.
This representation concerns the distribution
of response states and does not require
reproducing the token-level generation mechanism.

If the target transition probabilities are
continuous and strictly positive on the compact
simplex $\mathcal S$, their log-utilities are
continuous.
A utility family capable of uniformly
approximating continuous functions can therefore
approximate these log-utilities arbitrarily well.
The softmax map then transfers this approximation
to the transition probabilities.
Since the mean-field drift is a weighted sum
of these probabilities, for every $\varepsilon>0$
there exists a sufficiently expressive RUM
such that
\[
    \sup_{\boldsymbol\rho\in\mathcal S}
    \left\|
        \mathcal F_{\mathcal G}^{\mathrm{RUM}}
        (\boldsymbol\rho)
        -
        \mathcal F_{\mathcal G}
        (\boldsymbol\rho)
    \right\|_1
    <\varepsilon.
\]
This provides a conditional justification for
using RUM to approximate the population-level
dynamics induced by interacting AR LLMs.

\begin{theorem}[Mean-field limit and equilibrium accuracy]
\label{thm:qa_mean_field_equilibrium}
Consider the closed population process in which,
at each interaction round, one agent is selected
uniformly and transitions from state $a$ to state $b$
with probability
$\kappa_{ab}^{\mathrm{RUM}}
(\widehat{\boldsymbol\rho}_N^k;\mathcal G)$.
Suppose that
$\kappa_{ab}^{\mathrm{RUM}}
(\boldsymbol\rho;\mathcal G)\geq\epsilon$
for every $a,b\in\{H,D,T\}$ and
$\boldsymbol\rho\in\mathcal S$, where
$0<\epsilon\leq1/3$.
Assume further that, for some $0\leq L<3\epsilon$,
\begin{equation}
\begin{aligned}
    &\max_{a\in\{H,D,T\}}
    \sum_{b\in\{H,D,T\}}
    \left|
        \kappa_{ab}^{\mathrm{RUM}}
        (\boldsymbol\rho;\mathcal G)
        -
        \kappa_{ab}^{\mathrm{RUM}}
        (\widetilde{\boldsymbol\rho};\mathcal G)
    \right|
    \\
    &\qquad\leq
    L\|\boldsymbol\rho-\widetilde{\boldsymbol\rho}\|_1,
    \qquad
    \boldsymbol\rho,\widetilde{\boldsymbol\rho}
    \in\mathcal S.
\end{aligned}
\label{eq:qa_kernel_regularity}
\end{equation}
If
$\widehat{\boldsymbol\rho}_N^0
\xrightarrow{\Pr}\boldsymbol\rho^0\in\mathcal S$,
then the following statements hold.

\begin{enumerate}
    \item \textit{Mean-field limit.}
    For every finite $S>0$,
    \begin{equation}
        \sup_{0\leq s\leq S}
        \left\|
            \widehat{\boldsymbol\rho}_N^{\lfloor Ns\rfloor}
            -
            \boldsymbol\rho(s)
        \right\|_1
        \xrightarrow{\Pr}0,
        \qquad N\to\infty,
        \label{eq:qa_fluid_limit}
    \end{equation}
    where $\boldsymbol\rho(s)$ solves
    \[
        \frac{d\boldsymbol\rho(s)}{ds}
        =
        \mathcal F_{\mathcal G}^{\mathrm{RUM}}
        \bigl(\boldsymbol\rho(s)\bigr),
        \qquad
        \boldsymbol\rho(0)=\boldsymbol\rho^0.
    \]

    \item \textit{Unique attracting equilibrium.}
    The simplex $\mathcal S$ is positively invariant.
    The ODE has a unique equilibrium
    $\boldsymbol\rho^\star\in\mathcal S$,
    and every solution starting in $\mathcal S$
    converges to $\boldsymbol\rho^\star$.

    \item \textit{Equilibrium accuracy.}
    The equilibrium satisfies
    \[
        \rho_b^\star
        =
        \sum_{a\in\{H,D,T\}}
        \rho_a^\star
        \kappa_{ab}^{\mathrm{RUM}}
        (\boldsymbol\rho^\star;\mathcal G),
        \qquad b\in\{H,D,T\}.
    \]
    In particular,
    \begin{equation}
        \rho_T^\star
        =
        \frac{
            \rho_H^\star
            \kappa_{HT}^{\mathrm{RUM}}
            (\boldsymbol\rho^\star;\mathcal G)
            +
            \rho_D^\star
            \kappa_{DT}^{\mathrm{RUM}}
            (\boldsymbol\rho^\star;\mathcal G)
        }{
            \kappa_{TH}^{\mathrm{RUM}}
            (\boldsymbol\rho^\star;\mathcal G)
            +
            \kappa_{TD}^{\mathrm{RUM}}
            (\boldsymbol\rho^\star;\mathcal G)
        }.
        \label{eq:qa_equilibrium_accuracy}
    \end{equation}
\end{enumerate}
\end{theorem}

The proof is given in
Appendix~\ref{subsubsec:qa_equilibrium}.
The equilibrium determines the eventual proportions
of hallucinated, uncertain, and correct responses.
The result concerns the closed RUM state model;
its application to the full AR network depends
on the accuracy of the RUM approximation and
the mean-field closure. We note that open-source code for implementing this collaborative Q\&A can be found at \texttt{https://github.com/aditj/llm\_network\_tmlr}.

\subsection{Example 2. Structural Bias and Glass-Ceiling Effects}

In this section, we study structural bias and
glass-ceiling effects (GCE) in networks autonomously
formed by interacting LLM agents.

\subsubsection{Why Study Glass-Ceiling Effects?}

In social networks, glass-ceiling
effects describe barriers that limit certain groups' access to senior positions beyond
what differences in relevant qualifications can
explain.
Such disadvantages become more pronounced at higher
levels of the hierarchy and can intensify over
the course of a career \cite{cotter2001glass}.
In social networks, related effects manifest as
underrepresentation among highly connected nodes:
homophily, unequal group sizes, and preferential
attachment can jointly generate this disparity
\cite{avin2015homophily,nettasinghe2022scale},
while social recommendation algorithms can further
amplify it \cite{stoica2018algorithmic}.

These findings motivate studying whether analogous
barriers emerge in networks of interacting LLM agents.
Following \cite{zhang2026emergence}, we distinguish
two cases.
In the \emph{capability-aligned} case, stronger agents
become more influential, potentially improving
collective performance.
In the \emph{capability-misaligned} case, weaker agents
acquire disproportionate influence while stronger
agents remain peripheral, suppressing
useful contributions and amplifying errors.
The main questions are:
\begin{quote}
How do local interaction decisions shape network
influence? Can capable agent types remain
structurally disadvantaged in the long run?
\end{quote}


We construct a mean-field ODE for type-level
incoming and outgoing centrality.
By estimating its expected allocations from
LLM-agent data, the ODE provides a parsimonious
model for predicting network evolution.
Under suitable conditions,
the centrality dynamics converge to a unique
stable fixed point, which determines whether
an average GCE persists in the long run.

\subsubsection{Interacting AR Model to Mean Field Dynamics}

Unlike collaborative Q\&A in Example 1, the communication network now
evolves through agent-generated decisions.
Write $\mathcal G^k=(\mathcal V^k,W^k)$, where
each directed edge $(j,i)$ carries a weight
$w^k(j,i)\in\mathbb R_+^d$ recording accumulated
interaction effort, such as token exchange,
interaction frequency, and reasoning effort.
An absent edge has zero weight, and
$\mathcal N_i^k
=\{j\in\mathcal V^k:w^k(j,i)\neq\mathbf 0\}$
is agent $i$'s incoming neighbor set.

The agent $i$ activated at interaction round $k+1$
generates its response according to
\begin{equation}
\begin{gathered}
    x_t^i(k+1)
    \sim p_{\theta_i}\left(
        \cdot\mid c^i(k),x_{1:t-1}^i(k+1)
    \right),
    \qquad k\geq0,\\
    c^i(k)=
    \left\{
        c^i(0),x^i(k),
        \{x^j(k):j\in\mathcal N_i^k\}
    \right\},
    \qquad k\geq1.
\end{gathered}
\label{eq:gce_network_ar}
\end{equation}
Here, $c^i(0)$ is the fixed initial context,
and the context components are formatted into
an ordered token sequence.
The generated response determines network changes:
\begin{equation}
    \boldsymbol\Delta^{k+1}
    =\Gamma\left(x^i(k+1)\right),
    \qquad
    W^{k+1}=W^k+\boldsymbol\Delta^{k+1}.
    \label{eq:gce_network_evolution}
\end{equation}
The map $\Gamma$ extracts interaction partners
and nonnegative edge-weight contributions,
including contributions to previously absent edges.
Assuming positive total contributions in every
dimension, we normalize them so that
$\sum_{j,i}\boldsymbol\Delta^{k+1}(j,i)=\mathbf 1_d$.
Thus, the network shapes agents' contexts,
and their responses shape subsequent network changes.

\paragraph{Type-level centrality and average GCE.}

Let $\mathcal V_R^k$ and
$\mathcal V_B^k=\mathcal V^k\setminus\mathcal V_R^k$
denote the two agent types, with each agent's
type fixed throughout the process.
Starting from zero weights, the total network
weight after $k$ rounds is $k\mathbf 1_d$.
New agents, if introduced, have zero existing
edge weights.
Define the empirical type-$R$ centrality as
\begin{equation}
    \widehat{\boldsymbol\Theta}^k
    =
    \begin{pmatrix}
        \widehat{\boldsymbol\theta}_{\mathrm{in}}^k\\
        \widehat{\boldsymbol\theta}_{\mathrm{out}}^k
    \end{pmatrix}
    =
    \frac{1}{k}
    \sum_{i\in\mathcal V_R^k}
    \sum_{j\in\mathcal V^k}
    \begin{pmatrix}
        w^k(j,i)\\
        w^k(i,j)
    \end{pmatrix},
    \qquad k\geq1.
    \label{eq:gce_centrality_definition}
\end{equation}
Its components are the realized shares of
incoming and outgoing weight associated with
type $R$.
Write
$\boldsymbol\Theta^k
=\mathbb E[\widehat{\boldsymbol\Theta}^k]$;
both vectors belong to $\mathcal K=[0,1]^{2d}$.

Following \cite{zhang2026emergence}, define
type-level communication influence by the
outgoing-to-incoming weight ratios
\begin{equation}
\begin{aligned}
    I^k(R)
    &=
    \frac{\|\widehat{\boldsymbol\theta}_{\mathrm{out}}^k\|_1}
         {\|\widehat{\boldsymbol\theta}_{\mathrm{in}}^k\|_1},\quad
    I^k(B)=
    \frac{d-\|\widehat{\boldsymbol\theta}_{\mathrm{out}}^k\|_1}
         {d-\|\widehat{\boldsymbol\theta}_{\mathrm{in}}^k\|_1}.
\end{aligned}
\label{eq:gce_type_influence}
\end{equation}
These ratios measure information supplied to
other agents relative to information received.
To formalize the requirement that type $R$ has
substantially smaller long-run relative influence,
fix $\eta\in(0,1)$.
Type $R$ experiences an \emph{average GCE at level
$\eta$} if
\begin{equation}
    \limsup_{k\to\infty}
    \frac{I^k(R)}{I^k(B)}
    \leq\eta
    \qquad\text{almost surely},
    \label{eq:average_gce_definition}
\end{equation}
with the ratios well defined for all sufficiently
large $k$.
Smaller $\eta$ indicates a stronger disparity.
When type $R$ has comparable or greater
task capability, this criterion identifies
a persistent capability--influence mismatch.

\paragraph{Mean-field dynamics.}

Let $\boldsymbol\Delta_R^{k+1}$ collect the
incoming and outgoing weight added to type $R$
at round $k+1$.
Normalization by the total network weight gives
\begin{equation}
    \widehat{\boldsymbol\Theta}^{k+1}
    =
    \widehat{\boldsymbol\Theta}^k
    +
    \frac{1}{k+1}
    \left(
        \boldsymbol\Delta_R^{k+1}
        -\widehat{\boldsymbol\Theta}^k
    \right),
    \qquad k\geq1.
    \label{eq:gce_empirical_recursion}
\end{equation}
Assume that the conditional mean of the next
increment given the full history is determined
by $\mathcal G^k$ and depends on it only through
$\widehat{\boldsymbol\Theta}^k$:
\begin{equation}
    \mathbb E\!\left[
        \boldsymbol\Delta_R^{k+1}
        \mid\mathcal G^k
    \right]
    =
    \boldsymbol\mu(
        \widehat{\boldsymbol\Theta}^k
    ),
    \qquad
    \boldsymbol\mu:\mathcal K\to\mathcal K.
    \label{eq:gce_conditional_closure}
\end{equation}
Under this type-level closure, define
$\mathcal F_H(\boldsymbol\Theta)
=\boldsymbol\mu(\boldsymbol\Theta)-\boldsymbol\Theta$.
Taking expectations yields
\begin{equation}
    \boldsymbol\Theta^{k+1}
    =
    \boldsymbol\Theta^k
    +
    \frac{1}{k+1}
    \mathbb E\!\left[
        \mathcal F_H(
            \widehat{\boldsymbol\Theta}^k
        )
    \right].
    \label{eq:gce_centrality_update}
\end{equation}
On the rescaled time axis
$s_k=\sum_{\ell=1}^k1/\ell$, the associated
mean-field ODE is
\begin{equation}
    \frac{d\boldsymbol\Theta(s)}{ds}
    =
    \mathcal F_H\bigl(\boldsymbol\Theta(s)\bigr),
    \qquad s\in\mathbb R_+.
    \label{eq:gce_mean_field}
\end{equation}
Stochastic approximation theory relates the
empirical recursion to this ODE under 
suitable stochastic averaging theory
conditions.

\subsubsection{Cross-Attention Utility Model (CAM)}

Our final modeling step, before analyzing the
stability of the mean-field dynamics, is to
characterize how interacting LLM agents respond
to information received from other agents.
For this purpose, we use the cross-attention
mechanism of transformers
\cite{vaswani2017attention}.
Cross-attention provides a mechanism for modeling
how interacting LLM agents respond to information
supplied by other agents. A query represents the
information sought by the target, each source key
describes the information available from that source,
and the corresponding value encodes the content
to be aggregated. Query--key compatibility determines
how strongly each source value contributes to the
aggregate \cite{vaswani2017attention}.
This mechanism supports information integration
in perception and question answering
\cite{jaegle2021perceiver,izacard2021leveraging}.

The same structure motivates a model of
communication between LLM agents:
a target agent's query represents its current
information needs, while a source agent's key
and value represent the relevance and content
of its potential contribution, respectively.
Following \cite{zhang2026emergence}, the
cross-attention response utility model (CAM)
uses query--key compatibility to weight
source contributions.
Together with candidate-selection and
normalization rules, these weighted contributions
determine the modeled communication allocations,
linking agents' contexts to the evolution of
the interaction network.                  These allocations determine the mean-field drift:
\begin{equation}
    \mathcal F_H(\boldsymbol\Theta)
    =
    \begin{pmatrix}
        \boldsymbol\mu_{\mathrm{in}}(\boldsymbol\Theta)
        -\boldsymbol\theta_{\mathrm{in}}\\
        \boldsymbol\mu_{\mathrm{out}}(\boldsymbol\Theta)
        -\boldsymbol\theta_{\mathrm{out}}
    \end{pmatrix}.
    \label{eq:gce_drift_components}
\end{equation}
Each component compares the expected new allocation
with the corresponding current centrality share.
The CAM construction is detailed in Appendix~\ref{subsec:gce}.

\subsubsection{Stability and Average GCE at Equilibrium}

We now give sufficient conditions for centrality
convergence and connect the limiting allocation
to average GCE.

\begin{theorem}[Convergence of centrality]
\label{thm:gce_equilibrium}
Consider \eqref{eq:gce_empirical_recursion} on
$\mathcal K=[0,1]^{2d}$.
Assume that
$\widehat{\boldsymbol\Theta}^1\in\mathcal K$,
agent types remain fixed,
$\boldsymbol\Delta_R^{k+1}\in\mathcal K$,
and $\mathcal G^k$ determines the conditional
mean of the next increment given the full history.
Suppose that
\eqref{eq:gce_conditional_closure} holds and,
for some $0\leq L<1$,
\begin{equation}
\begin{aligned}
    \|\boldsymbol\mu(\boldsymbol\Theta)
      -\boldsymbol\mu(\widetilde{\boldsymbol\Theta})\|_2
    &\leq
    L\|\boldsymbol\Theta
         -\widetilde{\boldsymbol\Theta}\|_2,
    \\
    &\qquad
    \boldsymbol\Theta,\widetilde{\boldsymbol\Theta}
    \in\mathcal K.
\end{aligned}
\label{eq:gce_contraction_condition}
\end{equation}
Then:
\begin{enumerate}
    \item \textit{Unique attracting equilibrium.}
    The ODE \eqref{eq:gce_mean_field} has a unique
    equilibrium $\boldsymbol\Theta^\star\in\mathcal K$
    satisfying
    $\boldsymbol\mu(\boldsymbol\Theta^\star)
    =\boldsymbol\Theta^\star$.
    Every ODE trajectory starting in $\mathcal K$
    converges to $\boldsymbol\Theta^\star$.

    \item \textit{Centrality convergence.}
    The empirical centrality satisfies
    $\widehat{\boldsymbol\Theta}^k
    \to\boldsymbol\Theta^\star$ almost surely,
    and its expectation satisfies
    $\boldsymbol\Theta^k\to\boldsymbol\Theta^\star$.
\end{enumerate}
\end{theorem}

The proof is given in
Appendix~\ref{subsubsec:gce_equilibrium}.
The contraction condition limits feedback from
current prominence to future allocation.
The same result applies to the CAM surrogate
when its formation process satisfies the stated
closure and contraction assumptions.

To connect the fixed point to average GCE, suppose
\[
    0<\|\boldsymbol\theta_{\mathrm{in}}^\star\|_1<d,
    \qquad
    \|\boldsymbol\theta_{\mathrm{out}}^\star\|_1<d.
\]
Almost-sure convergence and
\eqref{eq:gce_type_influence} imply
\begin{equation}
    \frac{I^k(R)}{I^k(B)}
    \xrightarrow{\mathrm{a.s.}}
    \frac{
        \|\boldsymbol\theta_{\mathrm{out}}^\star\|_1
        \bigl(d-\|\boldsymbol\theta_{\mathrm{in}}^\star\|_1\bigr)
    }{
        \|\boldsymbol\theta_{\mathrm{in}}^\star\|_1
        \bigl(d-\|\boldsymbol\theta_{\mathrm{out}}^\star\|_1\bigr)
    }.
    \label{eq:gce_relative_influence_limit}
\end{equation}
Therefore, average GCE at level $\eta$ occurs
if and only if
\begin{equation}
    \frac{
        \|\boldsymbol\theta_{\mathrm{out}}^\star\|_1
        \bigl(d-\|\boldsymbol\theta_{\mathrm{in}}^\star\|_1\bigr)
    }{
        \|\boldsymbol\theta_{\mathrm{in}}^\star\|_1
        \bigl(d-\|\boldsymbol\theta_{\mathrm{out}}^\star\|_1\bigr)
    }
    \leq\eta.
    \label{eq:gce_equilibrium_criterion}
\end{equation}
In particular, a limiting relative influence
below one is equivalent to
$\|\boldsymbol\theta_{\mathrm{out}}^\star\|_1
<
\|\boldsymbol\theta_{\mathrm{in}}^\star\|_1$.
Thus, convergence establishes that influence
allocation stabilizes, while the fixed point
determines whether the allocation retains
an average GCE. We note that open-source code for implementing this GCE effect can be found at \texttt{https://github.com/yimingz1218/LLM\_GCE}.

{\bf Summary}.  
Parts 1 and 2 developed a dynamical-systems perspective
on LLMs, from individual nonlinear autoregressive models
to networks of interacting agents. In the large-population
limit, mean-field ODEs connect local stochastic interactions
to emergent network structure, including preferential
attachment and persistent structural bias manifested
through type-dependent glass-ceiling effects.

\section{Part 3. Flexible Time-Series Generative Modeling via Neural SDEs}

In this part of the paper,
we consider a complementary class  of  generative AI models:  neural stochastic differential equations (neural SDEs)~\cite{tzen2019neural,kidger2021nsdegan,tzen2019theoretical,li2020scalable}
for generating continuous-time stochastic time-series trajectories,
including those observed at irregular time intervals.
Our presentation is a high-level  exposition of  neural SDEs and their training methodology.
Technical details are deferred to the cited literature.

\paragraph*{Overview of Neural SDEs}
Whereas AR models generate sequences through successive conditional distributions, neural stochastic differential equations (SDEs) generate stochastic trajectories through learned drift and diffusion coefficients of an SDE. Both approaches specify probability laws on sequences or paths through parameterized stochastic dynamics. Neural SDEs extend the dynamical-systems perspective of generative AI to continuous-time processes, particularly for modeling irregularly sampled time series and other path-valued data.

Let us first explain the main ideas informally.
We express a  neural SDE in the It\^o form
as
\begin{equation}
\label{eq:sde1}
dX_t^\theta = \mu_{\theta}(t,X_t^\theta)\,dt + \sigma_{\theta}(t,X_t^\theta)\,dW_t.
\end{equation}
Here $W_t$ denotes standard Brownian motion, and \eqref{eq:sde1} has specified  initial condition $X_0^{\theta}$. We denote the path-law of \eqref{eq:sde1} as $P(\theta)$ -- this is the probability law on an appropriate function space. Also,  the drift $\mu_{\theta}$ and volatility $\sigma_{\theta}$ are parametrized as neural networks with weights $\theta$. 

A neural SDE is a generator in the same sense as a generative adversarial network or a score-based diffusion~\cite{song2021sde}: one samples noise and pushes it through a parametrized map to obtain draws from an unknown law~\cite{kidger2021nsdegan}. Here that law is a measure on \emph{path space} rather than on a Euclidean state space. This is the natural model for stochastic time series, including irregularly sampled observations, and it has been used for market-path simulation in
quantitative finance~\cite{choudhary2024funvol}, latent
dynamics of motion-capture trajectories~\cite{li2020scalable}, data-driven
turbulence closures~\cite{boral2023niles}, and reconstruction of noisy
gene-regulation series~\cite{zhang2025gene}. The usefulness of neural SDEs is captured in part by their \emph{universal approximation} of arbitrary stochastic dynamics \cite{kwossek2025universal}, inheriting the flexibility and expressivity that have made neural networks widely applicable function approximators.

\begin{figure*}[t]
  \centering
  \includegraphics[width=\textwidth]{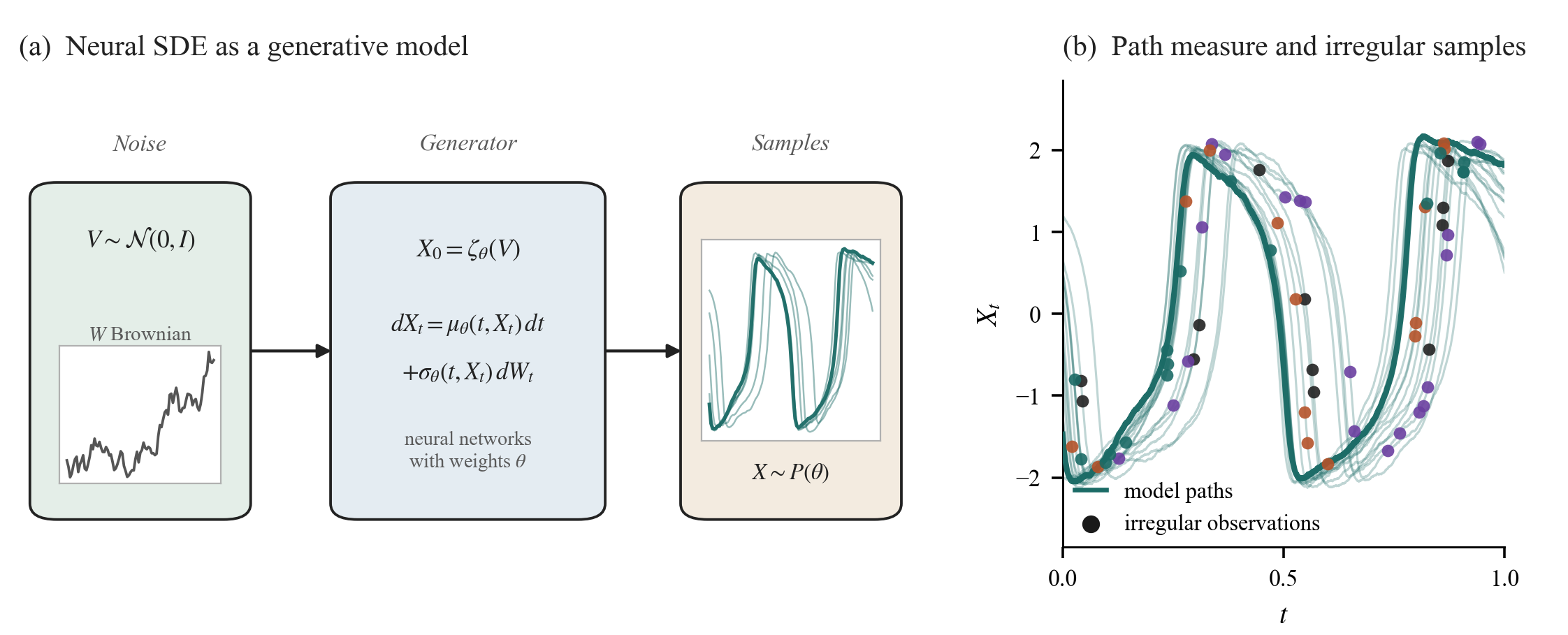}
  \caption{Neural SDEs as path-space generative models. (a)~Initial noise \(V\) and Brownian motion \(W\) are mapped through
  neural networks with weights \(\theta\) to sample paths \(X\sim P(\theta)=\mathcal{L}(X^{\theta})\).
  Sampling is a numerical SDE solve. (b)~The induced object is a path measure (thin trajectories). Observed data may be irregular samples on those trajectories, so the  natural training target is \(P(\theta)\), not a shared time grid or the terminal law \(\mathcal{L}(X_T)\).}
  \label{fig:nsde-pathspace}
\end{figure*}

Figure~\ref{fig:nsde-pathspace} illustrates this generative modeling pipeline. Panel~(a) is the generator: \(\zeta_\theta\) determines the law of \(X_0\), and \((\mu_\theta,\sigma_{\theta})\) are the parametrized drift and volatility functions of the neural SDE; learning \(\theta\) is then equivalent to learning the law of the dynamics if the neural parametrization is rich enough. Panel~(b) illustrates a collection of irregularly sampled data points, which do not determine a common discrete mesh or a single endpoint; however, they can be modeled as \emph{evaluations along continuous trajectories}. Section~\ref{sec:nsde} distinguishes matching the path law \(P(\theta)\) from matching the terminal-time distribution \(\mathcal{L}(X_T)\). 

\paragraph*{Main Ideas}
``Learning" \(\theta\) in general means making the path law \(P(\theta)\) close to an empirical measure which corresponds to \emph{observed data}, by minimizing an appropriate discrepancy functional which captures the distance between $P(\theta)$ and the data. Existing methods differ in the discrepancy they minimize, e.g., adversarial path-space objectives~\cite{kidger2021nsdegan}, variational free
energy~\cite{tzen2019neural}, finite-dimensional distribution matching~\cite{zhang2024efficient}, signature-kernel scores~\cite{issa2024sigker}, and in how gradients of that discrepancy are
computed, e.g.\ automatic differentiation through an SDE solver~\cite{tzen2019neural} or stochastic adjoints~\cite{kidger2021efficient}. Section~\ref{sec:nsde} organizes these choices; Appendix~\ref{sec:wsc} then presents the key idea of  Wiener-space cubature~\cite{lyons2004cubature,snow2025efficient} -- which serves as a useful alternative
 to Monte Carlo path simulation. 

The main results of this section are: 
\begin{enumerate}[(i)]
    \item We introduce the mathematical framework of neural SDEs, and motivate them as principled path-space generative models;
    \item We organize training methods by the law they match and by the differentiation engine/estimator pair they use, including representative time and memory scalings;
    \item We summarize Wiener-space cubature as a drop-in replacement for Monte Carlo estimation of path-functional objectives, following~\cite{snow2025efficient}.
\end{enumerate}

Before presenting the main neural SDE framework, we first introduce
the high level main results of our Wiener space cubature technique for improving the efficiency of neural SDE training by exploiting high-order deterministic approximations to SDE solutions.

\subsection{Wiener-Space Cubature for  Neural SDE Training}
\label{sec:wsc_intro}

Figure~\ref{fig:nsde-pathspace} displays the neural SDE generative modeling pipeline. The generative modeling optimization proceeds by first specifying a discrepancy between the path law $P(\theta)$ and data, then estimating and differentiating that discrepancy to optimize the generative law. The default evaluation method in most  estimators is via  Monte Carlo simulation of the SDE \eqref{eq:sde1} along independent driving Brownian motion paths, with root-MSE sampling error of $\CO(n^{-1/2})$.

Wiener-space cubature~\cite{lyons2004cubature,snow2025efficient} enables replacing 
the Brownian motion driving an SDE 
 with a finite collection of judiciously chosen deterministic trajectories, yielding an ordinary differential equation for each trajectory. 
 Brownian motion is replaced by a finite set of bounded-variation paths $\{\om_z\}$ with weights $\{\lambda_z\}$, chosen once and independently of $\theta$, which match the Brownian motion statistics up to high order; Appendix~\ref{sec:wsc} specifies the precise sense in which this is meant. Then, each neural SDE path evaluation becomes an ordinary differential equation along $\om_z$, and the training objective becomes a weighted sum of ODE trajectories. The key consequences are:
\begin{enumerate}[(i)]
    \item \emph{ODE gradients.} Brownian paths need not be stored, and the $\theta$-gradient $\nabla_\theta$ (a crucial step in backpropagation) can be evaluated via deterministic ODE adjoints rather than an SDE adjoint~\cite{chen2018neural, matsubara2021symplectic}.
    \item \emph{Fewer paths.} Under suitable conditions, the root-MSE in the number $n$ of ODE solves improves from $\CO(n^{-1/2})$ to $\CO(n^{-1})$~\cite{snow2025efficient}.
\end{enumerate}
These efficiency improvements are complementary: in reasonable settings we need fewer trajectories for a prescribed estimation accuracy, and each trajectory is cheaper since it is a deterministic ODE. These efficiency gains require suitable structure in the neural SDE drift and volatility functions and must be weighed against the preprocessing cost of computing the cubature paths and weights; see \cite{snow2025efficient}.

We make this more precise now. Write $\Xt\in\CX$ for a sample path of the neural SDE~\eqref{eq:nsde} (defined in Section~\ref{sec:nsde}), and let $\Xd$ be a fixed empirical path measure on the space $\CX$ of continuous trajectories on $[0,T]$. A data-dependent path functional
\[
\CLd
\;=\;
\CE_{X\sim\Xd}[\CL(\cdot,X)]:\CX\to\reals
\]
reduces matching the law $P(\theta)\approx\Xd$ to
\begin{equation}
\label{eq:objective}
\arg\min_{\theta\in\Theta\subset\reals^{d_\theta}}
\CE_{\Xt\sim P(\theta)}[\CLd(\Xt)].
\end{equation}
(We seek a local stationary point.)
Wiener space cubature is an estimator of~\eqref{eq:objective}; it is compatible with the discrepancies and differentiation engines of Section~\ref{sec:nsde}.

\subsubsection{Efficient Neural SDE Training Guarantees}
\label{sec:wsc_gr_intro}

The estimate and its gradient are as follows. Proofs and the full cubature path construction are in~\cite{snow2025efficient}; Appendix~\ref{sec:wsc} expands on this and provides the definitions used to make these objects precise.

\begin{mtheorem}[Wiener Space Cubature Approximation, Informal]
\label{thm:meta}
The objective~\eqref{eq:objective} admits the deterministic surrogate
\begin{equation}
\label{eq:mto}
\CE[\CLd(\Xt)]
\;=\;
\sum_{z=1}^{n}\lambda_z\,\CLd(\phisoln)
\;+\;
\text{error},
\end{equation}
where $\phisoln\in\CX$ is the solution on $[0,T]$ of the Stratonovich ODE with the vector fields of~\eqref{eq:sde1}, driven by a cubature path
$\om_z$. The weights satisfy $\lambda_z\ge 0$ and
$\sum_{z=1}^{n}\lambda_z=1$, and $(\lambda_z,\om_z)$ do not depend on
$\theta$. After suitable high-order recombination of cubature paths, we  obtain $\text{error}=\CO(n^{-1})$.
\end{mtheorem}

Evaluating the Wiener space cubature objective in \eqref{eq:mto} enables the minimization \eqref{eq:objective} with rate $\CO(n^{-1})$:
\begin{equation}
\label{eq:mineq}
\min_{\theta\in\Theta}
\sum_{z=1}^{n}\lambda_z\,\CLd(\phisoln)
\;=\;
\min_{\theta\in\Theta}\CE[\CLd(\Xt)]
\;+\;
\CO(n^{-1}).
\end{equation}

Furthermore, each path evaluation is now a deterministic ODE and we thus obtain improved efficiency of backpropagation, as introduced in the following Meta Corollary. 
\begin{mcorollary}[Efficient ODE Gradient Evaluations]
\label{cor:mcor}
Given~\eqref{eq:mto},
\begin{equation}
\label{eq:gradeval}
\frac{\del}{\del\theta}
\sum_{z=1}^{n}\lambda_z\,\CLd(\phisoln)
\;=\;
\sum_{z=1}^{n}\lambda_z
\frac{\del}{\del\theta}\CLd(\phisoln),
\end{equation}
where each summand is computed by automatic differentiation or an ODE
adjoint~\cite{chen2018neural,matsubara2021symplectic}.
These evaluations are typically cheaper in time and memory than
differentiation through SDE solvers~\cite{tzen2019neural,kidger2021efficient}.
\end{mcorollary}

\remark{\emph{Why more efficient than Monte Carlo?}}
Monte Carlo approximates \(\CE[\CLd(\Xt)]\) by an empirical average of \(n\) independent Brownian paths. That estimator does \emph{not} exploit the structure of the Wiener measure, so the root-mean-squared error is a central-limit fluctuation of size \(\CO(n^{-1/2})\).

Wiener space cubature treats the same expectation as a quadrature problem on path space. The quantities that enter the stochastic Taylor expansion of solutions of~\eqref{eq:sde1} are a finite list of iterated Stratonovich integrals. Tchakaloff's theorem~\cite{bayer2006proof} is a finite-dimensional analogue of this quadrature problem, and says that a
measure on \(\mathbb{R}^d\) with finite moments of degree \(m\) admits an \emph{exact} cubature rule for all polynomials of degree \(\le m\), with at
most \(\dim\mathcal{P}_m = \left( \genfrac{}{}{0pt}{}{d+m}{m}\right)\) (dimension of the vector space of polynomials on $\reals^d$ of total degree at most $m$) positive-weight nodes. In~\cite{lyons2004cubature} an analogous statement is made for the Wiener measure, with polynomials replaced by iterated integrals of degree
\(\le m\): there is a discrete path measure, supported on a number of bounded-variation paths only depending on $m$, that matches these iterated integrals exactly. The approximation error is then the unmatched remainder of the Taylor expansion (a deterministic bias) rather than a sampling variance. This, together with a particular high-order recombination scheme \cite{litterer2012high}, yields the
\(\CO(n^{-1})\) rate in Meta Theorem~\ref{thm:meta}.
Appendix~\ref{sec:wsc} makes this precise. 

However, these efficiency improvements hold only under suitably structured neural SDE drift and volatility functions, and after an appropriate pre-processing step has been completed which computes the cubature paths and weights. In-depth discussion of the trade-offs between these necessary computational contingencies and the possible efficiency gains can be found in \cite{snow2025efficient}. 

Going forward, Section~\ref{sec:nsde} introduces the neural SDE model and the training taxonomy in more detail.
Appendix~\ref{sec:wsc} then gives the Wiener space cubature construction behind
\eqref{eq:mto}--\eqref{eq:gradeval}.

We note that open-source code for implementing this Wiener space cubature approach to neural SDE training can be found at \texttt{https://github.com/LukeSnow0/Neural-SDE}.

\subsection{Neural SDEs: Generative Modeling and Training}
\label{sec:nsde}

We now specify the neural SDE model and the common training choices underlying discrepancy evaluations such as introduced in~\eqref{eq:objective}. 

\subsubsection{Model class}

We first define the neural SDE model more precisely. Fix a finite time horizon $T > 0$. Let
\[
W : [0, T] \to \mathbb{R}^{d_b},
\qquad
v \sim N(0, I_{d_v}),
\]
be a standard Brownian motion (the process $W$ in
Figure~\ref{fig:nsde-pathspace}) and an initial latent, respectively. A neural stochastic differential equation is given by, for $t\in[0,T]$,
\begin{equation}
X_0^\theta = \zeta_{\theta}(v),
\quad
dX_t^\theta = \mu_{\theta}(t,X_t^\theta)\,dt + \sigma_{\theta}(t,X_t^\theta)\,dW_t,
\label{eq:nsde}
\end{equation}
where
\begin{align*}
&\zeta_{\theta} : \mathbb{R}^{d_v} \to \mathbb{R}^{d_x},
\quad
\mu_{\theta} : [0,T]\times \mathbb{R}^{d_x} \to \mathbb{R}^{d_x}, \\
&
\sigma_{\theta} : [0,T]\times \mathbb{R}^{d_x} \to \mathbb{R}^{d_x\times d_b},
\end{align*}
are neural networks collectively parametrized by
$\theta \in \Theta \subset \mathbb{R}^{d_\theta}$.

Let $\CX$ denote the space of continuous paths on $[0,T]$. A sample path of \eqref{eq:nsde} is $\Xt := (X^{\theta}_t)_{t\in[0,T]}$, and the path-space law is
\begin{equation}
P(\theta):=\Law\big(\Xt\big).
\label{eq:path_law_sec2}
\end{equation}
This is the law $P(\theta)$ in Figure~\ref{fig:nsde-pathspace} and in \eqref{eq:objective}.

Used as a generative model, \eqref{eq:nsde} is trained to fit a data law either at a fixed terminal time $T$ or on the path space $\CX$.
Terminal-time training matches a finite-dimensional readout $\Law(h_\theta(X_T^\theta))$ and sits in the same family as score-based diffusions~\cite{song2021sde}, which learn a reverse drift for a \emph{prescribed} forward noising process rather than a fully learned pair $(\mu_\theta,\sigma_{\theta})$.
Path-space training matches $P(\theta)$ itself, i.e.\ infinite-dimensional generative modeling of a trajectory law~\cite{kidger2021nsdegan}. That is the natural target for stochastic time series, including irregularly sampled observations~\cite{rubanova2019latentode,li2020scalable}. Both regimes are developed in Section~\ref{sec:nsde-regimes}.

We next outline details for these distributional training objectives, as well as neural optimization techniques in each of these distinct training regimes.

\subsubsection{Distributional training regimes}
\label{sec:nsde-regimes}

Let \(X_{\mathrm{data}}\) denote the empirical data law on the path space \(\mathcal{X}\)
(or on a finite-dimensional observation space derived from it). Training a neural SDE
consists of choosing \(\theta\) so that the model law is close to this data law in the sense of a \emph{discrepancy} functional \(J(P(\theta),X_{\text{data}})\) measuring an appropriate distance between $P(\theta)$ and $X_{\text{data}}$. The choice of \(J\) is independent of how the resulting objective is estimated or differentiated, which is the subject of Section~\ref{sec:nsde-opt}. It is useful to distinguish path-space from terminal-time learning; we include finite-dimensional distribution (FDD) matching as the computationally important intermediate~\cite{zhang2024fdm}.

\paragraph{Path-space learning}
In path-space learning, the target is the law of the entire trajectory. One fits \(P(\theta)=\mathcal{L}(X^\theta)\)
to \(X_{\mathrm{data}}\) via
\begin{equation}
\label{eq:path-obj}
\theta^\star \;\in\;
\operatorname*{arg\,min}_{\theta\in\Theta}
J\bigl(P(\theta),\,X_{\mathrm{data}}\bigr).
\end{equation}
This is the natural regime for stochastic time-series generation, irregularly sampled
sequential data, and any setting in which temporal evolution itself is part of the modeling target. Even if a decoder is applied pathwise, \(Y_t=h_\theta(X^\theta_t)\), the problem remains one of matching laws on path space.

Here we list two representative discrepancies in order to make~\eqref{eq:path-obj} explicit. Integral probability metrics (IPM) with a neural path-functional critic~\cite{kidger2021nsdegan} take the form
\begin{equation}
\label{eq:ipm}
J_{\mathrm{IPM}}(P,Q)
:=
\sup_{f\in\mathcal{F}}
\bigl(\mathbb{E}_{X\sim P}[f(X)]-\mathbb{E}_{Y\sim Q}[f(Y)]\bigr),
\end{equation}
typically with \(\mathcal{F}\) parametrized by a neural controlled differential equation.
Signature-kernel scoring rules~\cite{issa2024sigker} replace the adversarial inner supremum here by a closed-form kernel score on \(\mathcal{X}\). In both cases the decision
variable is a \emph{path measure}, not a fixed finite-dimensional terminal distribution.

\paragraph{Finite-dimensional distribution matching}
An intermediate target (between path-law and terminal-time modeling) is the collection of finite-dimensional laws
\begin{align}
\begin{split}
\label{eq:fdd}
&\mu_{t_{1:k}}(\theta)
\;:=\;
\mathcal{L}\bigl(X^\theta_{t_1},\ldots,X^\theta_{t_k}\bigr),\\
&0\le t_1<\cdots<t_k\le T.
\end{split}
\end{align}
In this setting, one minimizes \(J(\mu_{t_{1:k}}(\theta),\mu^{\mathrm{data}}_{t_{1:k}})\) on a grid
(or a sum of such discrepancies over subgrids), rather than a discrepancy on the continuous path-space $\mathcal{X}$ directly. Because~\eqref{eq:nsde} is Markov, matching a generating family of FDDs, in particular consecutive pairs, given the initial law, is enough in principle to control the process, and reduces the objective-side cost from quadratic in
the number of time steps to linear~\cite{zhang2024fdm}. FDD matching is therefore path-valued in spirit, but finite-dimensional in the quantity actually compared; it should not be conflated with terminal-time learning, which uses only \(t=T\).

\paragraph{Terminal and latent-variable learning}
A distinct regime arises when the observations are finite-dimensional random variables
derived from the dynamics, most commonly the terminal state or a readout of it. Write
\(Z^\theta=h_\theta(X^\theta_T)\). Here, one matches \(\mathcal{L}(Z^\theta)\) to an empirical
distribution \(\nu_{\mathrm{data}}\) via
\begin{equation}
\label{eq:terminal-obj}
\theta^\star \;\in\;
\operatorname*{arg\,min}_{\theta\in\Theta}
J\bigl(\mathcal{L}(Z^\theta),\,\nu_{\mathrm{data}}\bigr).
\end{equation}
This is the natural viewpoint when the neural SDE is a continuous-time latent-variable generator rather than a direct model of an observed series. The standard realization is variational: an evidence lower bound on the observation likelihood, as in the diffusion-limit latent Gaussian models of Tzen and Raginsky~\cite{tzen2019neural} and the scalable latent SDEs of Li \emph{et al.}~\cite{li2020scalable}.

Score-based generative models~\cite{song2021sde} sit in this terminal-time family, but they
are a strict special case. There, the forward noising process is prescribed, the diffusion
coefficient is not learned, and the neural network parametrizes the score \(\nabla\log p_t\) appearing in the reverse drift. A general neural SDE instead learns both \(\mu_\theta\) and \(\sigma_{\theta}\) in~\eqref{eq:nsde}, and is thus more general in both the neural parametrization and the target distribution (possibly path-dependent) which is matched.

\subsubsection{Optimization of the training objective}
\label{sec:nsde-opt}

Section~\ref{sec:nsde-regimes} specified \emph{what} is matched. Training still requires
efficient evaluation of \(\nabla_\theta J_\theta\), where \(J_\theta\) denotes the chosen discrepancy as a function of the neural-SDE parameters $\theta$. There are two largely orthogonal design choices:
\begin{enumerate}[(i)]
\item a \emph{pathwise differentiation engine}, which approximately evaluates \(\nabla_\theta\) of a
functional (discrepancy) of a single realization of~\eqref{eq:nsde};
\item an \emph{objective / expectation estimator}, which approximates \(J_\theta\)
(typically an expectation under \(P(\theta)\), or a saddle-point in the adversarial case)
by averaging, critic updates, or a deterministic surrogate.
\end{enumerate}
A training method consists of these choices, of which we list several below. Differentiation engine costs below are quoted per simulated path of \(L\) solver steps; estimator costs multiply or add to that baseline, and here \(D\) denotes the state dimension of \(X^\theta\).

\begin{table*}[!t]
\centering
\caption{Representative complexity of neural SDE training.
Engine costs are \emph{per path} of \(L\) solver steps; \(D\) is the state dimension.
Estimator costs multiply or add to the chosen engine.
\(M\) is the number of simulated (variational) paths, \(K\) the number of critic steps
per generator step, and \(n\) the number of cubature ODEs, vs. \(n\) Monte Carlo paths in a standard expectation evaluation.}
\label{tab:nsde-complexity}
\begin{tabular}{llll}
\hline
Method & Time & Memory & Comment \\
\hline
\multicolumn{4}{l}{\textit{Pathwise differentiation engines (per path)}}\\
AD through solver
  & \(\mathcal{O}(L)\) & \(\mathcal{O}(L)\) & exact relative to discretization \\
Stochastic adjoint
  & \(\mathcal{O}(L)\) & \(\mathcal{O}(1)\) in \(L\) & default for long trajectories \\
Forward sensitivity
  & \(\mathcal{O}(LD)\) & \(\mathcal{O}(1)\) in \(L\) & poor in high dimension \\
\hline
\multicolumn{4}{l}{\textit{Objective / expectation estimators}}\\
Variational MC
  & engine \(\times M\) & engine \(\times M\) & ELBO; \(M\) SDE paths \\
Adversarial
  & engine \(\times K\) & critic + engine & \(K\) critic steps \\
Signature-kernel score
  & engine \(+\,\mathcal{O}(L^2)\) & objective-dependent & kernel / PDE overhead \\
FDM
  & engine \(+\,\mathcal{O}(L)\) & objective-dependent & Markov reduction \\
Cubature / ODE surrogate
  & \(n\) ODE solves & ODE-style & no Brownian MC; rate \(\mathcal{O}(n^{-1})\) \\
\hline
\end{tabular}
\end{table*}

\paragraph{Pathwise differentiation engines}
\begin{itemize}
\item[-] \textit{Automatic differentiation.}
Simulate a discretized neural SDE and backpropagate through the solver graph,
\(\nabla_\theta J(X^\theta)\)~\cite{tzen2019neural}.
Time and memory are both \(\mathcal{O}(L)\) per path. This is exact relative to the chosen discretization and is simplest for short trajectories.

\item[-] \textit{Stochastic adjoint backpropagation.}
Rather than storing the forward graph, solve a backward adjoint SDE associated with the
loss, reconstructing  the primal path~\cite{li2020scalable,kidger2021efficient}.
Time remains \(\mathcal{O}(L)\) per path, while memory drops to \(\mathcal{O}(1)\) in \(L\).
This is often the default method in settings where \(L\) is large.

\item[-] \textit{Forward sensitivity propagation.}
Propagate Jacobian information \(\partial X_t/\partial\theta\) along the forward solve.
The solver graph need not be stored, but the state is enlarged by the sensitivity, giving
time \(\mathcal{O}(LD)\) and memory \(\mathcal{O}(1)\) in \(L\) \cite{ma2021comparison}. This scaling is unattractive
once the dimension $D$ is moderate.
\end{itemize}

\paragraph{Objective / expectation estimators}
\begin{itemize}
\item[-] \textit{Variational (reparametrized) Monte Carlo.}
In latent-SDE models one maximises an evidence lower bound
\begin{align}
\begin{split}
\label{eq:elbo}
\log p_\theta(y_{1:K})
\;&\ge\;
\mathbb{E}_{q_\phi}\bigl[\log p_\theta(y_{1:K}\mid X_{0:T})\bigr]
\\&\qquad -\mathrm{KL}\bigl(q_\phi\,\Vert\,P(\theta)\bigr),
\end{split}
\end{align}
using \(M\) simulated paths of the variational process~\cite{tzen2019neural,li2020scalable}. Here $p_{\theta}$ and $q_\phi$ are appropriately parametrized variational probability distributions. The total cost is then that of the chosen engine, times \(M\).

\item[-] \textit{Adversarial objectives.}
For an integral probability
metric~\eqref{eq:ipm} one alternates critic and generator updates~\cite{kidger2021nsdegan}.
With \(K\) critic steps per generator step the training cost is multiplied by roughly
\(K\). 

\item[-] \textit{Score- and kernel-based objectives.}
Signature-kernel scores incur an objective-side cost \(\mathcal{O}(L^2)\) in the number of
discretization steps~\cite{issa2024sigker}. Finite-dimensional distribution matching
reduces this to \(\mathcal{O}(L)\) by exploiting the Markov structure of~\eqref{eq:nsde} 
\cite{zhang2024fdm}. Both are estimators of a discrepancy from
Section~\ref{sec:nsde-regimes}; they still require a differentiation engine.

\item[-] \textit{Deterministic expectation approximations.}
Wiener space cubature, previewed in Section~\ref{sec:wsc_intro}, replaces Monte Carlo by the weighted ODE ensemble of~\eqref{eq:mto},
\begin{equation}
\label{eq:cubature-estimator}
J_\theta
\;\approx\;
\sum_{z=1}^{n}\lambda_z\,\CLd(\phisoln),
\end{equation}
with $\lambda_z\ge 0$, $\sum_z\lambda_z=1$, and each $\phisoln$ driven by a bounded-variation cubature path~\cite{lyons2004cubature}. Gradients then pass through ODE solvers (Corollary~\ref{cor:mcor}). Under the conditions of~\cite{snow2025efficient}, the weak error improves
from \(\mathcal{O}(n^{-1/2})\) to \(\mathcal{O}(n^{-1})\).
\end{itemize}

Table~\ref{tab:nsde-complexity} summarizes the resulting time and memory scalings.
The two axes are independent: adjoint + signature kernel, AD + cubature, and so on, are all well-defined pairs. Implementation details for the first seven rows are in the cited papers. 

Monte Carlo estimation converges at
\(\mathcal{O}(n^{-1/2})\) and requires simulating and storing Brownian paths. Wiener-space cubature, developed in Appendix~\ref{sec:wsc}, keeps the same pathwise loss \(L_{\mathrm{data}}\) but replaces
that estimator by~\eqref{eq:cubature-estimator}: fewer trajectories, each of them an ODE, differentiated with standard ODE adjoints.

\subsection{Discussion}
\label{sec:nsde-close}

This section introduced neural SDEs as path-space generative models: noise \((v,W)\) is pushed through neural vector fields~\eqref{eq:nsde} to a law \(P(\theta)\) on \(\CX\), which is the natural target when observations are irregular trajectories, rather than a shared grid or a terminal snapshot (Figure~\ref{fig:nsde-pathspace}).
Training of neural SDEs is comprised by two largely independent choices. Section~\ref{sec:nsde-regimes} specified \emph{what} is matched (the path
measure, a generating family of finite-dimensional distributions, or a terminal/latent readout).
Section~\ref{sec:nsde-opt} specified \emph{how} the resulting discrepancy is estimated and differentiated (SDE adjoints, variational Monte Carlo, adversarial critics, signature kernels, or a deterministic cubature ensemble).
A training method is a pair of these choices.

The Wiener space cubature estimator replaces Brownian Monte Carlo by a discrete path measure that is exact for iterated Stratonovich integrals up to a chosen degree, i.e., Tchakaloff's theorem on Wiener space~\cite{lyons2004cubature}.
Each model solve is then an ODE, gradients are ODE adjoints (Corollary~\ref{cor:mcor}), and after recombination the weak error in the number of paths can improve from \(\CO(n^{-1/2})\) to
\(\CO(n^{-1})\) (Meta Theorem~\ref{thm:meta}).
The construction is a pre-processing step, independent of \(\theta\); the price is the need for an explicit degree-\(m\) rule (currently \(m=3,5,7\)) and a recombination of the concatenated support. Proofs, the recombination algorithm, and numerical comparisons with Monte Carlo training are in~\cite{snow2025efficient}. 
Importantly, cubature changes how training expectations
are computed, not the stochastic nature of the
generative model. A trained neural SDE still generates
random trajectories according to a learned probability
law on path space. Together with the autoregressive
LLMs and interacting networks of Parts 1 and 2,
this illustrates a common principle of generative AI:
learning stochastic dynamical models from data and
generating new realizations from the learned models.

\section{Conclusion}

This paper developed a stochastic dynamical-systems
perspective on three classes of generative AI models.
For LLMs, the nonlinear autoregressive formulation
connects pretraining by conditional maximum likelihood
to stochastic token generation. For networks of
interacting LLMs, mean-field dynamics characterize
how local stochastic interactions give rise to
emergent network structure, preferential attachment,
and type-dependent glass-ceiling effects.
For continuous-time generative modeling, neural SDEs
define probability distributions over stochastic
trajectories, with rough-path cubature providing
an alternative to Monte Carlo training.
Together, these examples illustrate how concepts
from  stochastic processes
and dynamical systems provide  insight
into the structure, collective behavior, and
computation of modern generative AI models.

\bibliographystyle{ieeetr}
\bibliography{Bibliography.bib}

\clearpage
\appendices

{\Large{\bf Supplementary Material}}

\section{Appendix for Part 2}

\subsection{Mean-Field Derivation and Proof of Theorem III.1}
\label{subsec:collaborative_qa}

We explain how the interacting autoregressive model
yields the mean-field dynamics in
\eqref{eq:qa_mean_field}.
We first approximate the state-transition
probabilities using a random utility model (RUM),
then derive the population ODE under a mean-field
closure and an asynchronous interaction schedule.
Finally, we prove
Theorem~\ref{thm:qa_mean_field_equilibrium}.

Recall that $x^i(k)$ denotes agent $i$'s current
response, $c^i(k)$ denotes its context, and
$\psi(x^i(k))\in\{H,D,T\}$ denotes its response state.
The empirical population vector is
$\widehat{\boldsymbol\rho}^k$, its expectation is
$\boldsymbol\rho^k
=\mathbb E[\widehat{\boldsymbol\rho}^k]$,
and the population state space is
\[
    \mathcal S
    =
    \left\{
        \boldsymbol\rho\in\mathbb R_+^3:
        \rho_H+\rho_D+\rho_T=1
    \right\}.
\]

\subsubsection{Construction of the RUM Transition Model}
\label{subsubsec:qa_rum}

We approximate the state-transition probabilities
induced by \eqref{eq:qa_network_ar} using the RUM
in \cite{jain2025collaborative}.
Let $a=\psi(x^i(k))$ denote agent $i$'s
current state.
For each possible next state $b\in\{H,D,T\}$,
let $\boldsymbol\varphi_b(c^i(k),a)$ be a feature
vector encoding the current state, neighbor count
$|\mathcal N_i|$, neighbors' response-state
composition, and textual context.
The deterministic utility is parameterized as
\[
    u_b(c^i(k),a;\phi)
    =
    \boldsymbol\alpha^\top
    \boldsymbol\varphi_b(c^i(k),a),
\]
where $\phi$ collects the fitted parameters,
including $\boldsymbol\alpha$ and any trainable
parameters of the feature representation.
The model assigns a random utility to each
candidate state and selects the state with
the largest utility:
\begin{equation}
\begin{aligned}
    r_b(c^i(k),a;\phi)
    &=
    u_b(c^i(k),a;\phi)+\varepsilon_b,
    \qquad
    \varepsilon_b
    \overset{\mathrm{i.i.d.}}{\sim}
    \operatorname{Gumbel}(0,1),
    \\
    \kappa_{ab}^{\mathrm{RUM}}(c^i(k);\phi)
    &:=
    \Pr_{\boldsymbol\varepsilon}\!\left(
        b=\operatorname*{arg\,max}_{z\in\{H,D,T\}}
        r_z(c^i(k),a;\phi)
    \right)
    \\
    &=
    \frac{
        \exp\!\left(u_b(c^i(k),a;\phi)\right)
    }{
        \sum_{z\in\{H,D,T\}}
        \exp\!\left(u_z(c^i(k),a;\phi)\right)
    }.
\end{aligned}
\label{eq:qa_rum_kernel}
\end{equation}
Any incentive or inference budget is held fixed
and suppressed in the notation.
Averaging over the Gumbel perturbations yields
the softmax transition probabilities.

The parameters are estimated by maximizing
the conditional log-likelihood of observed
state transitions, with $u_D(c,a;\phi)=0$
fixing the reference score.
Ground-truth state labels are used for fitting
and analysis, not supplied to the interacting agents.
The state-level model assumes Markov dependence
and homogeneous parameters within each
in-degree class.
Its probabilities approximate the AR-induced
state transitions; they do not specify the
full token distribution.

\subsubsection{Derivation of the Population Dynamics}
\label{subsubsec:qa_mean_field_derivation}

We derive the population drift by averaging
the state-transition probabilities over the
contexts encountered during interaction.
This requires a mean-field closure: conditional
on an agent's current state and the population
proportions, we approximate the distribution of
its context by a time-independent family of
distributions determined by the fixed network
and evidence allocation.

As in the overview,
$\kappa_{ab}(\boldsymbol\rho;\mathcal G)$
denotes the AR-induced transition probability
under this closure.
Its RUM approximation is
\begin{equation}
    \kappa_{ab}^{\mathrm{RUM}}
    (\boldsymbol\rho;\mathcal G,\phi)
    =
    \mathbb E_{\mathcal G,\boldsymbol\rho}
    \left[
        \kappa_{ab}^{\mathrm{RUM}}(c^i(k);\phi)
        \,\middle|\,
        \psi(x^i(k))=a
    \right],
    \qquad a,b\in\{H,D,T\}.
    \label{eq:qa_averaged_transition}
\end{equation}
The expectation averages over agents and the
contexts associated with their current state,
including the responses of all incoming neighbors.
Since each RUM transition distribution sums
to one,
$\sum_b\kappa_{ab}^{\mathrm{RUM}}
(\boldsymbol\rho;\mathcal G,\phi)=1$.
The fitted parameters $\phi$ remain fixed;
the dependence on $\boldsymbol\rho$ arises
from the changing distribution of contexts.
For readability, we suppress $\phi$ below.

The closure assumes that these averaged
probabilities can be described using
$\boldsymbol\rho$ alone.
When degree, capability, or evidence heterogeneity
must be retained, separate population fractions
are required for the corresponding agent classes,
as in the degree-based formulation of
\cite{jain2025collaborative}.

We use the asynchronous interaction schedule
introduced in the overview.
At each interaction round, one agent is sampled
uniformly and generates a new response using
the current responses of all its incoming neighbors.
All other agents retain their responses.
The empirical population fractions are
\begin{equation}
    \widehat\rho_b^k
    =
    \frac{1}{|\mathcal V|}
    \sum_{i\in\mathcal V}
    \mathbf 1\!\left\{
        \psi(x^i(k))=b
    \right\},
    \qquad b\in\{H,D,T\}.
    \label{eq:qa_empirical_fractions}
\end{equation}
Their expectations are
$\rho_b^k=\mathbb E[\widehat\rho_b^k]$,
while $\boldsymbol\rho(s)$ denotes the
deterministic large-population limit.

Under the closed RUM state model, the sampled
agent is in state $a$ with probability
$\widehat\rho_a^k$ and enters state $b$
with probability
$\kappa_{ab}^{\mathrm{RUM}}
(\widehat{\boldsymbol\rho}^k;\mathcal G)$.
A transition from $a$ to $b$ changes the population
vector by $(\mathbf e_b-\mathbf e_a)/|\mathcal V|$,
where $\mathbf e_b$ is the unit vector for state $b$.
Consequently,
\begin{equation}
\begin{aligned}
    &\mathbb E\!\left[
        \widehat\rho_b^{k+1}
        -
        \widehat\rho_b^k
        \,\middle|\,
        \widehat{\boldsymbol\rho}^k
    \right]
    \\
    &\qquad=
    \frac{1}{|\mathcal V|}
    \left[
        \sum_{a\in\{H,D,T\}}
        \widehat\rho_a^k
        \kappa_{ab}^{\mathrm{RUM}}
        (\widehat{\boldsymbol\rho}^k;\mathcal G)
        -
        \widehat\rho_b^k
    \right].
\end{aligned}
\label{eq:qa_population_increment}
\end{equation}
The first term counts the expected contribution
of the new response to state $b$; the second
removes the sampled agent's previous contribution.

The corresponding drift component is
\begin{equation}
\begin{aligned}
    \mathcal F_{\mathcal G,b}^{\mathrm{RUM}}
    (\boldsymbol\rho)
    &=
    \sum_{a\in\{H,D,T\}}
    \rho_a
    \kappa_{ab}^{\mathrm{RUM}}
    (\boldsymbol\rho;\mathcal G)
    -\rho_b
    \\
    &=
    \sum_{\substack{a\in\{H,D,T\}\\a\neq b}}
    \left[
        \rho_a
        \kappa_{ab}^{\mathrm{RUM}}
        (\boldsymbol\rho;\mathcal G)
        -
        \rho_b
        \kappa_{ba}^{\mathrm{RUM}}
        (\boldsymbol\rho;\mathcal G)
    \right].
\end{aligned}
\label{eq:qa_explicit_mean_field_drift}
\end{equation}
Collecting these components gives
$\mathcal F_{\mathcal G}^{\mathrm{RUM}}
=(\mathcal F_{\mathcal G,H}^{\mathrm{RUM}},
  \mathcal F_{\mathcal G,D}^{\mathrm{RUM}},
  \mathcal F_{\mathcal G,T}^{\mathrm{RUM}})^\top$.
Each component is the net population flow
into the corresponding state.
For state $T$, incoming flow represents correction
of hallucinated responses or resolution of
uncertainty, while outgoing flow represents
the loss of correct answers.

Taking expectations in
\eqref{eq:qa_population_increment} gives
\begin{equation}
    \boldsymbol\rho^{k+1}
    =
    \boldsymbol\rho^k
    +
    \frac{1}{|\mathcal V|}
    \mathbb E\!\left[
        \mathcal F_{\mathcal G}^{\mathrm{RUM}}
        (\widehat{\boldsymbol\rho}^k)
    \right].
    \label{eq:qa_rum_expected_recursion}
\end{equation}
Rescaling interaction time as
$s=k/|\mathcal V|$ yields the associated ODE
\begin{equation}
    \frac{d\boldsymbol\rho(s)}{ds}
    =
    \mathcal F_{\mathcal G}^{\mathrm{RUM}}
    \bigl(\boldsymbol\rho(s)\bigr).
    \label{eq:qa_rum_mean_field}
\end{equation}
Thus, replacing the AR-induced transition
probabilities by their RUM approximations
replaces $\mathcal F_{\mathcal G}$ in
\eqref{eq:qa_mean_field} by
$\mathcal F_{\mathcal G}^{\mathrm{RUM}}$.
Here and below, $\boldsymbol\rho(s)$ denotes
the limiting trajectory of the closed RUM model.

Each interaction changes the empirical fractions
by $O(1/N)$, where $N=|\mathcal V|$, and a bounded
interval of rescaled time contains $O(N)$ interactions.
The proof below shows that the accumulated
sampling fluctuations vanish as $N\to\infty$.
The asynchronous schedule is essential:
simultaneous generation by all agents instead
produces a discrete-time mean-field recursion.

\subsubsection{Proof of Theorem~\ref{thm:qa_mean_field_equilibrium}}
\label{subsubsec:qa_equilibrium}

We write $\widehat{\boldsymbol\rho}_N^k$
for the empirical population vector with $N$ agents.
For the large-population limit, we consider
closed population models governed by the same
averaged transition kernel
$\kappa_{ab}^{\mathrm{RUM}}
(\boldsymbol\rho;\mathcal G)$.

\begin{proof}
For the fixed network $\mathcal G$ and fitted
RUM parameters, define the probability matrix
\[
    K(\boldsymbol\rho)
    =
    \left[
        \kappa_{ab}^{\mathrm{RUM}}
        (\boldsymbol\rho;\mathcal G)
    \right]_{a,b\in\{H,D,T\}}
\]
and the population map
$\mathcal T(\boldsymbol\rho)
=K(\boldsymbol\rho)^\top\boldsymbol\rho$.
By \eqref{eq:qa_explicit_mean_field_drift},
\[
    \mathcal F_{\mathcal G}^{\mathrm{RUM}}
    (\boldsymbol\rho)
    =
    \mathcal T(\boldsymbol\rho)-\boldsymbol\rho.
\]
Row-stochasticity and
\eqref{eq:qa_kernel_regularity} imply that this
drift is Lipschitz on $\mathcal S$.

First, row-stochasticity gives
$\sum_b\mathcal F_{\mathcal G,b}^{\mathrm{RUM}}
(\boldsymbol\rho)=0$.
At a boundary point with $\rho_b=0$,
\[
    \mathcal F_{\mathcal G,b}^{\mathrm{RUM}}
    (\boldsymbol\rho)
    =
    \sum_{a\in\{H,D,T\}}
    \rho_a
    \kappa_{ab}^{\mathrm{RUM}}
    (\boldsymbol\rho;\mathcal G)
    \geq0.
\]
Thus the total population mass is preserved
and no component can become negative.
This proves positive invariance of $\mathcal S$,
and Lipschitz continuity gives a unique ODE
solution for every initial condition in $\mathcal S$.

We next prove the mean-field limit.
Subtracting the conditional expected increment
from each population increment produces a
martingale difference under the closed
Markov state model.
Each such difference has norm bounded by
a constant times $1/N$.
Over $\lfloor NS\rfloor$ interaction rounds,
the accumulated martingale $\boldsymbol M_N$
therefore satisfies
\[
    \mathbb E\!\left[
        \sup_{0\leq s\leq S}
        \|\boldsymbol M_N(s)\|_1^2
    \right]
    \leq \frac{C_S}{N}
\]
by coordinatewise application of Doob's inequality.
Writing
$\widehat{\boldsymbol\rho}_N(s)
=\widehat{\boldsymbol\rho}_N^{\lfloor Ns\rfloor}$,
the population recursion becomes
\begin{equation}
\begin{aligned}
    \widehat{\boldsymbol\rho}_N(s)
    &=
    \widehat{\boldsymbol\rho}_N(0)
    +
    \int_0^s
    \mathcal F_{\mathcal G}^{\mathrm{RUM}}
    (\widehat{\boldsymbol\rho}_N(u))\,du
    \\
    &\quad+
    \boldsymbol M_N(s)
    +
    \boldsymbol R_N(s).
\end{aligned}
\label{eq:qa_martingale_decomposition}
\end{equation}
Since the drift is the difference of two
probability vectors,
$\|\mathcal F_{\mathcal G}^{\mathrm{RUM}}
(\boldsymbol\rho)\|_1\leq2$.
The discretization remainder therefore satisfies
$\sup_{s\leq S}\|\boldsymbol R_N(s)\|_1\leq2/N$.
Comparing \eqref{eq:qa_martingale_decomposition}
with the integral form of the ODE and applying
Gronwall's inequality gives
\[
\begin{aligned}
    &\sup_{0\leq s\leq S}
    \|\widehat{\boldsymbol\rho}_N(s)
      -\boldsymbol\rho(s)\|_1
    \\
    &\quad\leq
    e^{C_F S}
    \left(
        \|\widehat{\boldsymbol\rho}_N(0)
          -\boldsymbol\rho^0\|_1
        +
        \sup_{0\leq s\leq S}
        \|\boldsymbol M_N(s)\|_1
        +
        \frac2N
    \right),
\end{aligned}
\]
where $C_F$ is a Lipschitz constant of the drift.
The initial-condition assumption and the
martingale bound prove \eqref{eq:qa_fluid_limit}.

We now establish contraction of $\mathcal T$.
Let $\mathbf1$ be the three-dimensional
all-ones vector.
Since every entry of $K(\boldsymbol\rho)$ is
at least $\epsilon$, the matrix
$K(\boldsymbol\rho)-\epsilon\mathbf1\mathbf1^\top$
is nonnegative and each of its rows sums to
$1-3\epsilon$.
For any vector $\boldsymbol v$ with
$\mathbf1^\top\boldsymbol v=0$, it follows that
\begin{equation}
\begin{aligned}
    \|K(\boldsymbol\rho)^\top\boldsymbol v\|_1
    &=
    \|(
        K(\boldsymbol\rho)
        -\epsilon\mathbf1\mathbf1^\top
    )^\top\boldsymbol v\|_1
    \\
    &\leq
    (1-3\epsilon)\|\boldsymbol v\|_1.
\end{aligned}
\label{eq:qa_kernel_mixing_bound}
\end{equation}
Decomposing the difference between two population
maps gives
\[
\begin{aligned}
    \mathcal T(\boldsymbol\rho)
    -
    \mathcal T(\widetilde{\boldsymbol\rho})
    &=
    K(\boldsymbol\rho)^\top
    (\boldsymbol\rho-\widetilde{\boldsymbol\rho})
    \\
    &\quad+
    [K(\boldsymbol\rho)-K(\widetilde{\boldsymbol\rho})]^\top
    \widetilde{\boldsymbol\rho}.
\end{aligned}
\]
Applying \eqref{eq:qa_kernel_mixing_bound} to the
first term and \eqref{eq:qa_kernel_regularity}
to the second yields
\begin{equation}
    \|\mathcal T(\boldsymbol\rho)
      -\mathcal T(\widetilde{\boldsymbol\rho})\|_1
    \leq
    (1-3\epsilon+L)
    \|\boldsymbol\rho-\widetilde{\boldsymbol\rho}\|_1.
    \label{eq:qa_population_contraction}
\end{equation}
Since $L<3\epsilon$, $\mathcal T$ is a contraction
of $\mathcal S$ into itself.
The Banach fixed-point theorem therefore gives
a unique fixed point $\boldsymbol\rho^\star$,
which is also the unique equilibrium of the ODE.

To establish global attraction, set
$d(s)=\|\boldsymbol\rho(s)-\boldsymbol\rho^\star\|_1$.
For a small $h>0$, the ODE gives
\[
\begin{aligned}
    \boldsymbol\rho(s+h)-\boldsymbol\rho^\star
    &=
    (1-h)(\boldsymbol\rho(s)-\boldsymbol\rho^\star)
    \\
    &\quad+
    h\bigl[
        \mathcal T(\boldsymbol\rho(s))
        -
        \mathcal T(\boldsymbol\rho^\star)
    \bigr]
    +o(h).
\end{aligned}
\]
Using \eqref{eq:qa_population_contraction},
the upper right Dini derivative satisfies
$D^+d(s)\leq-(3\epsilon-L)d(s)$.
Hence
$d(s)\leq e^{-(3\epsilon-L)s}d(0)$,
proving global convergence.

Finally, the fixed-point identity gives
$\rho_b^\star
=\sum_a\rho_a^\star
\kappa_{ab}^{\mathrm{RUM}}
(\boldsymbol\rho^\star;\mathcal G)$.
Setting the $T$ component of
\eqref{eq:qa_explicit_mean_field_drift} to zero
balances the flows into and out of the correct
response state and yields
\eqref{eq:qa_equilibrium_accuracy}.
The denominator is positive because each
transition probability is at least $\epsilon$.
\end{proof}

The assumptions separate state mixing from
social sensitivity.
The lower bound $\epsilon$ prevents any response
state from becoming inaccessible, while $L$
measures how strongly the transition probabilities
change with the population.
The condition $L<3\epsilon$ requires this
population feedback to be sufficiently weak
relative to state mixing.
Although finite RUM scores give positive
choice probabilities, a uniform lower bound
and the required sensitivity bound must still
be verified over the modeled contexts.

Equation~\eqref{eq:qa_equilibrium_accuracy}
is an implicit equilibrium relation, since
the transition probabilities are evaluated
at $\boldsymbol\rho^\star$.
Its numerator measures correction of hallucinated
and uncertain responses, while its denominator
measures the probability of leaving the correct
response state.
The theorem does not imply
$\rho_T^\star>\rho_T(0)$:
a stable population can still have low accuracy.
Its conclusions concern the closed RUM model;
application to the full AR network requires
justification of the approximation and closure.

\subsection{Mean-Field Derivation and Proof of Theorem III.2 }
\label{subsec:gce}

We connect response-driven network formation
to the centrality dynamics in
\eqref{eq:gce_mean_field}.
We introduce a cross-attention model (CAM)
for edge-weight proposals, derive the stochastic
centrality recursion, and prove
Theorem~\ref{thm:gce_equilibrium}.

The evolving network is
$\mathcal G^k=(\mathcal V^k,W^k)$,
with vector-valued edge weights $w^k(j,i)$.
The response-derived network increment is
$\boldsymbol\Delta^{k+1}$, and
$\boldsymbol\Delta_R^{k+1}$ collects its
type-$R$ incoming and outgoing contributions.
The empirical centrality is
$\widehat{\boldsymbol\Theta}^k$,
its expectation is
$\boldsymbol\Theta^k
=\mathbb E[\widehat{\boldsymbol\Theta}^k]$,
and $\boldsymbol\Theta(s)$ denotes the
associated deterministic ODE trajectory.

\subsubsection{Construction of the CAM Allocation Model}
\label{subsubsec:gce_CAM}

We approximate the edge-weight proposals underlying
$\Gamma$ in \eqref{eq:gce_network_evolution}
using a cross-attention-inspired utility model,
following \cite{zhang2026emergence}.
Let $\tau_i\in\{R,B\}$ denote agent $i$'s type,
and let
$\mathbf h_i^k=\operatorname{Enc}(c^i(k))$
be a latent representation of its context.
For a candidate interaction $j\to i$, define
\begin{equation}
    \mathbf U_{j\to i}^{k}
    =
    \left[
        \frac{
            (Q_{\tau_i}\mathbf h_i^k)^\top
            (K_{\tau_j}\mathbf h_j^k)
        }{\sqrt{d_a}}
        V_{\tau_j}\mathbf h_j^k
    \right]_+.
    \label{eq:gce_CAM_utility}
\end{equation}
The query $Q_{\tau_i}\mathbf h_i^k$ represents
the receiving agent's information needs,
the key $K_{\tau_j}\mathbf h_j^k$ represents
the candidate source's relevance, and
the value $V_{\tau_j}\mathbf h_j^k$ specifies
a vector of proposed edge-weight contributions.
Here, $d_a$ is the query--key dimension,
and $[\cdot]_+$ applies coordinatewise
nonnegative truncation.

Let $\phi$ collect the encoder and projection
parameters.
These parameters are fitted by minimizing
prediction error against edge-weight proposals
extracted from observed LLM responses.
Once fitted, they remain fixed during the
modeled dynamics.
We suppress their dependence in
$\mathbf h_i^k$ and $\mathbf U_{j\to i}^k$
for readability.

Each formation step contains candidate
interactions $(j_m,i_m)$,
$m=1,\ldots,M^{k+1}$, sampled according to
the formation protocol.
Candidates may include previously unconnected
agents, allowing new edges to form.
Newly introduced agents have zero existing
edge weights, and their context representations
are computed from their initial contexts.

To convert utilities into increments, we normalize
them separately in each edge-weight dimension.
The CAM approximation to the response-derived
increment is
\begin{equation}
    \boldsymbol\Delta^{k+1}(j,i)
    \approx
    \left[
        \sum_{m=1}^{M^{k+1}}
        \mathbf1\{(j_m,i_m)=(j,i)\}
        \mathbf U_{j_m\to i_m}^{k}
    \right]
    \oslash
    \left[
        \sum_{m=1}^{M^{k+1}}
        \mathbf U_{j_m\to i_m}^{k}
    \right],
    \label{eq:gce_normalized_increment}
\end{equation}
where $\oslash$ denotes coordinatewise division.
We assume that the denominator is positive
in every dimension.
The normalized surrogate increments sum to
$\mathbf1_d$, matching the normalization of
$\boldsymbol\Delta^{k+1}$ in the overview.
When these surrogate increments define the
modeled formation process,
\eqref{eq:gce_normalized_increment}
holds with equality.

The surrogate depends on both the learned
utilities and the candidate-selection protocol.
In particular, the expectation of normalized
utilities must be computed after normalization;
it generally differs from normalizing
their expectations.

\subsubsection{Derivation of the Centrality Dynamics}
\label{subsubsec:gce_dynamics}

We derive the centrality drift by tracking how
each normalized formation step changes the
type-$R$ edge-weight totals.
Let $\mathcal V_R^k$ denote the type-$R$ agents
at step $k$, with types fixed after assignment.
The type-$R$ incoming and outgoing weight
added at step $k+1$ is
\begin{equation}
    \boldsymbol\Delta_R^{k+1}
    =
    \sum_{i\in\mathcal V_R^{k+1}}
    \sum_{j\in\mathcal V^{k+1}}
    \begin{pmatrix}
        \boldsymbol\Delta^{k+1}(j,i)\\
        \boldsymbol\Delta^{k+1}(i,j)
    \end{pmatrix}
    \in[0,1]^{2d}.
    \label{eq:gce_type_increment}
\end{equation}
The first block records new weight received
by type-$R$ agents, and the second records
new weight originating from them.
When agents arrive, existing weights are extended
by zeros, so their arrival does not change
previously accumulated totals.

Starting from zero weights, the total network
weight after $k$ formation steps is
$k\mathbf1_d$.
Using \eqref{eq:gce_centrality_definition},
we obtain the exact recursion
\begin{equation}
\begin{aligned}
    \widehat{\boldsymbol\Theta}^{k+1}
    &=
    \frac{
        k\widehat{\boldsymbol\Theta}^k
        +\boldsymbol\Delta_R^{k+1}
    }{k+1}
    \\
    &=
    \widehat{\boldsymbol\Theta}^k
    +
    \frac{1}{k+1}
    \left(
        \boldsymbol\Delta_R^{k+1}
        -
        \widehat{\boldsymbol\Theta}^k
    \right),
    \qquad k\geq1.
\end{aligned}
\label{eq:gce_centrality_recursion}
\end{equation}
The subtraction arises because centrality is
normalized by a total weight that increases
at every step.

Let $\mathscr H_k$ denote the sigma-algebra
generated by the complete history through step $k$,
including the network, contexts, and past random
choices.
We impose a type-level mean-field closure:
the conditional expectation of the next increment
depends on this history only through
$\widehat{\boldsymbol\Theta}^k$.
Collect the expected allocations into
\begin{equation}
    \boldsymbol\mu(\boldsymbol\Theta)
    =
    \begin{pmatrix}
        \boldsymbol\mu_{\mathrm{in}}(\boldsymbol\Theta)\\
        \boldsymbol\mu_{\mathrm{out}}(\boldsymbol\Theta)
    \end{pmatrix},
    \qquad
    \boldsymbol\Theta
    =
    \begin{pmatrix}
        \boldsymbol\theta_{\mathrm{in}}\\
        \boldsymbol\theta_{\mathrm{out}}
    \end{pmatrix}.
    \label{eq:gce_expected_allocation}
\end{equation}
The closure and the centrality drift satisfy
\begin{equation}
\begin{aligned}
    \mathbb E[
        \boldsymbol\Delta_R^{k+1}
        \mid\mathscr H_k
    ]
    &=
    \boldsymbol\mu(
        \widehat{\boldsymbol\Theta}^k
    ),
    \\
    \mathcal F_H(\boldsymbol\Theta)
    &=
    \boldsymbol\mu(\boldsymbol\Theta)
    -
    \boldsymbol\Theta.
\end{aligned}
\label{eq:gce_allocation_drift}
\end{equation}
If the closure is stated conditional on
$\mathcal G^k$, the current network must also
be sufficient to determine this full-history
conditional mean.
If additional context or network statistics
remain relevant, they must be retained in
the state description.

The CAM approximates these expected allocations
through the normalized trial utilities.
Writing
\[
    \widetilde{\mathbf U}_m^k
    =
    \mathbf U_{j_m\to i_m}^k
    \oslash
    \sum_{n=1}^{M^{k+1}}
        \mathbf U_{j_n\to i_n}^k,
\]
define
\begin{equation}
\begin{aligned}
    \boldsymbol\mu_\phi(\boldsymbol\Theta)
    &:=
    \begin{pmatrix}
        \boldsymbol\mu_{\mathrm{in},\phi}
        (\boldsymbol\Theta)\\
        \boldsymbol\mu_{\mathrm{out},\phi}
        (\boldsymbol\Theta)
    \end{pmatrix}
    \\
    &=
    \mathbb E\!\left[
        \sum_{m=1}^{M^{k+1}}
        \begin{pmatrix}
            \mathbf1\{\tau_{i_m}=R\}
            \widetilde{\mathbf U}_m^k\\
            \mathbf1\{\tau_{j_m}=R\}
            \widetilde{\mathbf U}_m^k
        \end{pmatrix}
        \,\middle|\,
        \widehat{\boldsymbol\Theta}^k
        =\boldsymbol\Theta
    \right].
\end{aligned}
\label{eq:gce_CAM_allocation}
\end{equation}
The expectation averages over candidate selection
and the context distribution specified by
the closure.
Thus, CAM predicts local weight proposals,
while the formation protocol determines their
contributions to type-level centrality.

The induced approximate drift is
\begin{equation}
    \mathcal F_{H,\phi}(\boldsymbol\Theta)
    =
    \boldsymbol\mu_\phi(\boldsymbol\Theta)
    -
    \boldsymbol\Theta.
    \label{eq:gce_CAM_drift}
\end{equation}
Consequently,
\[
    \mathcal F_{H,\phi}(\boldsymbol\Theta)
    -
    \mathcal F_H(\boldsymbol\Theta)
    =
    \boldsymbol\mu_\phi(\boldsymbol\Theta)
    -
    \boldsymbol\mu(\boldsymbol\Theta).
\]
This is the allocation-to-drift relationship
used in the approximation statement
in the overview.

Returning to the empirical centrality process,
define the centered increment
\[
    \boldsymbol M^{k+1}
    =
    \boldsymbol\Delta_R^{k+1}
    -
    \boldsymbol\mu(
        \widehat{\boldsymbol\Theta}^k
    ).
\]
It satisfies
$\mathbb E[\boldsymbol M^{k+1}\mid\mathscr H_k]=0$,
and \eqref{eq:gce_centrality_recursion} becomes
\begin{equation}
    \widehat{\boldsymbol\Theta}^{k+1}
    =
    \widehat{\boldsymbol\Theta}^k
    +
    \frac{1}{k+1}
    \left[
        \mathcal F_H(
            \widehat{\boldsymbol\Theta}^k
        )
        +
        \boldsymbol M^{k+1}
    \right].
    \label{eq:gce_stochastic_approximation}
\end{equation}
Taking expectations recovers
\eqref{eq:gce_centrality_update}.

This is a stochastic approximation recursion
with step size $1/(k+1)$.
Its associated continuous time is
$s_k=\sum_{\ell=1}^{k}1/\ell$, which grows
approximately as $\log k$.
On this time scale, the associated ODE is
\eqref{eq:gce_mean_field}.
Replacing $\boldsymbol\mu$ by
$\boldsymbol\mu_\phi$ yields the surrogate ODE
with drift $\mathcal F_{H,\phi}$.
Unlike the fixed-population scaling in
Collaborative Q\&A, this diminishing step size
follows directly from normalization by
accumulated network weight.

\subsubsection{Proof of Theorem~\ref{thm:gce_equilibrium}}
\label{subsubsec:gce_equilibrium}

Let $\mathcal K=[0,1]^{2d}$.
By construction,
$\boldsymbol\mu:\mathcal K\to\mathcal K$.
To quantify average influence disparity,
consider the outgoing-to-incoming weight ratios
\begin{equation}
\begin{aligned}
    I_R(\boldsymbol\Theta)
    &=
    \frac{\|\boldsymbol\theta_{\mathrm{out}}\|_1}
         {\|\boldsymbol\theta_{\mathrm{in}}\|_1},
    \\
    I_B(\boldsymbol\Theta)
    &=
    \frac{d-\|\boldsymbol\theta_{\mathrm{out}}\|_1}
         {d-\|\boldsymbol\theta_{\mathrm{in}}\|_1},
    \\
    g(\boldsymbol\Theta)
    &=
    I_B(\boldsymbol\Theta)-I_R(\boldsymbol\Theta).
\end{aligned}
\label{eq:gce_average_influence}
\end{equation}
The ratios are defined when both types have
positive incoming weight.
A positive realized gap
$g(\widehat{\boldsymbol\Theta}^k)$ means that
type $R$ generates less outgoing influence
per unit of incoming weight than type $B$.
When type $R$ has comparable or greater
independently assessed capability, a persistent
positive gap provides an average GCE criterion.

\begin{proof}
Since $\boldsymbol\mu$ maps the complete set
$\mathcal K$ into itself and is a contraction,
the Banach fixed-point theorem gives a unique
$\boldsymbol\Theta^\star$ satisfying
$\boldsymbol\mu(\boldsymbol\Theta^\star)
=\boldsymbol\Theta^\star$.
By \eqref{eq:gce_allocation_drift},
this is exactly the equilibrium condition
$\mathcal F_H(\boldsymbol\Theta^\star)=0$.

The ODE preserves $\mathcal K$.
At a boundary point with coordinate
$\Theta_\ell=0$, the corresponding drift
is $\mu_\ell(\boldsymbol\Theta)\geq0$.
At a boundary point with $\Theta_\ell=1$,
the drift is
$\mu_\ell(\boldsymbol\Theta)-1\leq0$.
Thus the vector field points into
$\mathcal K$ at its boundary.

Let
$V(\boldsymbol\Theta)
=\|\boldsymbol\Theta-\boldsymbol\Theta^\star\|_2^2$.
Using the equilibrium identity,
Cauchy--Schwarz, and
\eqref{eq:gce_contraction_condition}, we obtain
\begin{equation}
\begin{aligned}
    \dot V
    &=
    2\left\langle
        \boldsymbol\Theta-\boldsymbol\Theta^\star,
        \boldsymbol\mu(\boldsymbol\Theta)
        -\boldsymbol\mu(\boldsymbol\Theta^\star)
        -(\boldsymbol\Theta-\boldsymbol\Theta^\star)
    \right\rangle
    \\
    &\leq
    -2(1-L)V.
\end{aligned}
\label{eq:gce_lyapunov_bound}
\end{equation}
Therefore every ODE trajectory starting in
$\mathcal K$ converges to
$\boldsymbol\Theta^\star$.

For the stochastic process, write
$\gamma_k=1/(k+1)$ and
$V_k=\|\widehat{\boldsymbol\Theta}^k
-\boldsymbol\Theta^\star\|_2^2$.
The recursion
\eqref{eq:gce_centrality_recursion} is a convex
combination of two vectors in $\mathcal K$,
so the empirical centrality remains in $\mathcal K$.
Expanding the square, taking conditional
expectations, and using
\eqref{eq:gce_allocation_drift} gives
\begin{equation}
    \mathbb E[V_{k+1}\mid\mathscr H_k]
    \leq
    V_k-2(1-L)\gamma_k V_k
    +2d\,\gamma_k^2.
    \label{eq:gce_stochastic_lyapunov}
\end{equation}
The final term follows from
$\|\boldsymbol\Delta_R^{k+1}
-\widehat{\boldsymbol\Theta}^k\|_2^2\leq2d$,
since both vectors belong to $\mathcal K$.

Because $\sum_k\gamma_k=\infty$ and
$\sum_k\gamma_k^2<\infty$,
the nonnegative almost-supermartingale
convergence lemma implies that $V_k$
converges almost surely and
$\sum_k\gamma_kV_k<\infty$.
The divergence of $\sum_k\gamma_k$ then
forces the limit of $V_k$ to be zero.
This proves
$\widehat{\boldsymbol\Theta}^k
\to\boldsymbol\Theta^\star$
almost surely.

Since $\widehat{\boldsymbol\Theta}^k$
takes values in the bounded set $\mathcal K$,
dominated convergence also gives
\[
    \boldsymbol\Theta^k
    =
    \mathbb E[
        \widehat{\boldsymbol\Theta}^k
    ]
    \longrightarrow
    \boldsymbol\Theta^\star.
\]

Finally, the assumption
$0<\|\boldsymbol\theta_{\mathrm{in}}^\star\|_1<d$
makes both influence ratios well-defined
and continuous in a neighborhood of the equilibrium.
Almost-sure centrality convergence therefore
makes the realized gap well-defined for all
sufficiently large $k$ and gives
$g(\widehat{\boldsymbol\Theta}^k)
\to g(\boldsymbol\Theta^\star)$ almost surely.
Writing
$A=\|\boldsymbol\theta_{\mathrm{in}}^\star\|_1$
and
$B=\|\boldsymbol\theta_{\mathrm{out}}^\star\|_1$,
we have
\[
    g(\boldsymbol\Theta^\star)
    =
    \frac{d-B}{d-A}-\frac BA
    =
    \frac{d(A-B)}{A(d-A)}.
\]
The denominator is positive, so the sign
of the limiting gap equals the sign of
$A-B$, proving
\eqref{eq:gce_equilibrium_criterion}.
If this gap is strictly positive,
convergence implies that the realized gap
is eventually positive almost surely.
\end{proof}

The contraction condition limits the strength
of feedback from current prominence to future
edge allocation.
For the CAM surrogate, the same theorem applies
with $\boldsymbol\mu_\phi$ and
$\mathcal F_{H,\phi}$ in place of
$\boldsymbol\mu$ and $\mathcal F_H$,
provided that its formation process satisfies
the stated closure and contraction assumptions.
This condition concerns the combined CAM,
normalization rule, and candidate-selection
protocol; it does not follow from the
cross-attention formula alone.

The exact and surrogate equilibria satisfy
$\boldsymbol\mu(\boldsymbol\Theta^\star)
=\boldsymbol\Theta^\star$ and
$\boldsymbol\mu_\phi(\boldsymbol\Theta_\phi^\star)
=\boldsymbol\Theta_\phi^\star$, respectively.
Convergence of the fitted surrogate does not
by itself establish convergence of the original
AR-driven network, which requires the
corresponding assumptions for its conditional
increments.

The average influence criterion differs from
representation relative to population size.
If type $R$ has limiting population share $r$,
its outgoing representation can also be compared
with $r\mathbf1_d$.
Neither benchmark is guaranteed by convergence:
a unique stable equilibrium may retain
a substantial influence disparity.

\section{Appendix for Part 3. Wiener Space Cubature for Neural SDE Training}

\label{sec:wsc}

The aim of this Wiener space cubature method is to provide a finite set of deterministic paths and weights which approximate the statistics of Brownian motion, and which can be exploited to obtain high-order approximations of expected discrepancy evaluations in the neural SDE training context.

Meta Theorem~\ref{thm:meta} and Corollary~\ref{cor:mcor} used a cubature ensemble $\{\om_z,\lambda_z\}$ and ODE trajectories $\phisoln$.
We now define those objects and state the path functional bound that justifies~\eqref{eq:mto}.
Proofs are in~\cite{snow2025efficient}.

\subsection{Preliminaries. Cubature on Wiener space.}
\label{sec:wsc_prelim}

Cubature on Wiener space~\cite{lyons2004cubature} is a deterministic replacement for the law of $W$ in~\eqref{eq:nsde}: one matches all iterated
Stratonovich integrals up to a chosen degree, then realizes the diffusion as an ODE along those paths.
The underlying expansion is the weak Taylor formula
of~\cite{kloeden1992numerical}.

We first write~\eqref{eq:nsde} in Stratonovich form: we absorb time-dependence of $(\mu_\theta,\sigma_{\theta})$ by the usual augmentation $x\leftarrow(t,x)$, and apply the It\^o--Stratonovich correction, so that there are $\bdim+1$ autonomous fields $V_0^\theta,\ldots,V_{\bdim}^\theta$ and $\bdim$ Brownian components:
\begin{equation}
\mathrm{d}X_t^\theta
=
V_0^\theta(X_t^\theta)\,\mathrm{d}t
+
\sum_{i=1}^{\bdim} V_i^\theta(X_t^\theta)\circ\mathrm{d}W_t^i,
\qquad
X_0^\theta=x_0,
\label{eq:strat_sde}
\end{equation}
with the convention $\circ\mathrm{d}W_t^0:=\mathrm{d}t$. Here $V_i^\theta(\,\cdot\,)=\sigma_{\theta}(t,\,\cdot\,)_i$ for $i\ge 1$, and $V_0^\theta$ is the corrected drift. Throughout this subsection the initial condition $x_0$ is deterministic; random $X_0^\theta=\zeta_{\theta}(v)$ is restored in Section~\ref{sec:cpc} by conditioning.

Let $\mathcal A_m$ be the index set
\begin{equation}
\label{eq:CAm}
\mathcal A_m
=
\bigl\{
(i_1,\ldots,i_\ell)\in\{0,\ldots,d_b\}^\ell:
\ell+\mathrm{card}\{j:i_j=0\}\le m
\bigr\}.
\end{equation}
For $\psi\in C_b^\infty(\reals^{d_x};\reals)$, the weak Taylor formula gives, at time $\tau>0$,
\begin{equation}
\label{eq:sol_exp}
\mathbb{E}[\psi(X_\tau^\theta)]
=
\sum_{(i_1,\ldots,i_\ell)\in\mathcal A_m}
\bigl(V^\theta_{i_1:i_\ell}\psi\bigr)(x_0)
\cdot
\mathbb{E}
\Bigl[
\int_{0<s_1<\cdots<s_\ell<\tau}
\circ\mathrm{d}W^{i_1}_{s_1}\cdots\circ\mathrm{d}W^{i_\ell}_{s_\ell}
\Bigr]
+
O\bigl(\tau^{(m+1)/2}\bigr).
\end{equation}
Each $V_i^\theta$ acts as the directional derivative $(V_i^\theta\psi)(x):=\langle\nabla\psi(x),V_i^\theta(x)\rangle$, and \\
$V^\theta_{i_1:i_\ell}\psi
:=V^\theta_{i_1}(V^\theta_{i_2}(\cdots(V^\theta_{i_\ell}\psi)\cdots))$.
The times $s_1<\cdots<s_\ell$ parametrize the simplex of an iterated integral of length $\ell\le m$; they are not a numerical time grid.

\begin{definition}[Cubature Paths and Weights~\cite{lyons2004cubature}]
\label{def:cub}
Let $C^{0}_{\mathrm{bv}}([0,\tau];\mathbb{R}^{\bdim})$ denote continuous
paths of bounded variation on $[0,\tau]$.
A collection $\om_1,\ldots,\om_{\wcn}$ in that space, together with weights
$\lambda_1,\ldots,\lambda_{\wcn}>0$ such that
$\sum_{j=1}^{\wcn}\lambda_j=1$, defines a cubature formula of degree $m$
at time $\tau$ if, for every $(i_1,\ldots,i_\ell)\in\mathcal A_m$,
\begin{equation}
\label{eq:itr_eq}
\mathbb{E}
\Bigl[
\int_{0<s_1<\cdots<s_\ell<\tau}
\circ\mathrm{d}W^{i_1}_{s_1}\cdots\circ\mathrm{d}W^{i_\ell}_{s_\ell}
\Bigr]
=
\sum_{j=1}^{\wcn}\lambda_j
\int_{0<s_1<\cdots<s_\ell<\tau}
\mathrm{d}\om^{i_1}_j(s_1)\cdots\mathrm{d}\om^{i_\ell}_j(s_\ell).
\end{equation}
\end{definition}

For any $m\in\mathbb{N}$ and $\wct\ge 0$ there exist
$\wcn\le\mathrm{card}(\mathcal{A}_m)$ such paths and
weights~\cite{lyons2004cubature}.
Explicit rules are known for degrees $3$ and $5$~\cite{lyons2004cubature}
and degree $7$~\cite{ferrucci2026high}.
The practical consequence is that $\mathbb{E}[\psi(X_\wct^\theta)]$ can be
evaluated by a \emph{weighted ensemble of ODEs}.

\begin{assumption}[Smoothness, polynomial growth]
\label{as:smooth}
For a fixed cubature degree $m$, the fields
$V_0^\theta,\ldots,V_{\bdim}^\theta$ in~\eqref{eq:strat_sde} lie in
$C^{m+1}(\mathbb{R}^{\xdim};\mathbb{R}^{\xdim})$ and are globally Lipschitz
in $x$.
Moreover, for every multi-index $\alpha$ with $|\alpha|\le m+1$ there exist
$C_\alpha,\rho_\alpha>0$ such that
$\|\partial^\alpha V_i^\theta(x)\|
\le C_\alpha\bigl(1+\|x\|^{\rho_\alpha}\bigr)$ for
$i=0,\ldots,\bdim$.
\end{assumption}

\begin{proposition}[Cubature ODE approximation~\cite{lyons2004cubature}]
\label{prop:cub_ode}
Let $\Phi_{\wct,x_0}(\om)$ be the solution at time $\wct$ of
\begin{equation}
\label{eq:cub_ode}
\mathrm{d}\widetilde{x}_t
=
\sum_{i=0}^{\bdim} V_i^\theta(\widetilde{x}_t)\,\mathrm{d}\om^i(t),
\qquad
\widetilde{x}_0=x_0,
\end{equation}
with $\mathrm{d}\om^0(t):=\mathrm{d}t$.
Then, under Assumption~\ref{as:smooth}, for every
$\psi\in C_b^\infty(\reals^{\xdim};\reals)$,
\begin{equation}
\label{eq:cub_approx}
\Bigl|
\mathbb{E}[\psi(X_{\wct}^\theta)]
-
\sum_{j=1}^{\wcn}\lambda_j\,\psi\bigl(\Phi_{\wct,x_0}(\om_j)\bigr)
\Bigr|
=
O\bigl(\wct^{(m+1)/2}\bigr).
\end{equation}
\end{proposition}

Proposition~\ref{prop:cub_ode} is Taylor expansion of~\eqref{eq:cub_ode} together with the moment matching~\eqref{eq:itr_eq}: the ODE and the SDE
share the same iterated integrals up to degree $m$.
The path $\phisoln$ in~\eqref{eq:mto} is this ODE trajectory, run over $[0,T]$ rather than a single time $\wct$.

\textbf{Concatenation and recombination.}
\label{sec:recom}
The bound~\eqref{eq:cub_approx} is local, on an interval of length $\tau$. To reduce it one partitions $[0,\tau]$ into $N_\tau$ pieces
$0=\tau_0<\cdots<\tau_{N_\tau}=\tau$ and concatenates the same degree-$m$ formula on each piece~\cite[Prop.~3.2]{lyons2004cubature}. This partition is a numerical grid on the integration interval; it is distinct from the iterated-integral simplex in~\eqref{eq:itr_eq} and from any observation grid used to evaluate $\CLd$. The concatenated collection has $\wcn^{N_\tau}$ paths.
Writing $\delta_n:=\tau_n-\tau_{n-1}$, the weak error along the concatenated ensemble is
\begin{equation}
\label{eq:recom_err}
O\Biggl(
\delta_{N_\tau}^{1/2}
+
\sum_{n=1}^{N_\tau-1}
\delta_n^{(m+1)/2}\,(\tau-\tau_n)^{-m/2}
\Biggr).
\end{equation}
High-order recombination~\cite{litterer2012high} then thins this exponential support to polynomial cardinality while preserving the matched moments.
That thinning is what converts Theorem~\ref{thm:cubap} into the
$\CO(n^{-1})$ claim of Meta Theorem~\ref{thm:meta}; the algorithm is in \cite{snow2025efficient}.

\subsection{Path-functional cubature}
\label{sec:prelim}

Proposition~\ref{prop:cub_ode} controls $\mathbb{E}[\psi(X_T^\theta)]$ for a smooth test function $\psi$ of the \emph{state at a single time}.
However, the loss $\CLd$ in~\eqref{eq:objective} is a functional of the whole trajectory. The step from pointwise cubature to~\eqref{eq:mto} is a  Wasserstein bound for Lipschitz path functionals~\cite{snow2025efficient}.

Let us introduce the following technical assumptions which are necessary for the remaining quantitative contributions.
\begin{myassumptions}
\item[\nl{as:vecbd}{A1}]
The maps $\zeta_{\theta},\mu_{\theta},\sigma_{\theta}$ in~\eqref{eq:nsde}
are Lipschitz in the state, $\CE_v[\|\zeta_{\theta}(v)\|^2]<\infty$, and there
exists $M<\infty$ such that
\[
\sup_{t\in[0,T],\,x\in\reals^{\xdim},\,\theta\in\Theta}
\Bigl\{
\|\mu_{\theta}(t,x)\|
+\bigl\|\tfrac{\del}{\del x}\mu_{\theta}(t,x)\bigr\|
+\|\sigma_{\theta}(t,x)\|
+\bigl\|\tfrac{\del}{\del x}\sigma_{\theta}(t,x)\bigr\|
\Bigr\}
\le M.
\]

\item[\nl{as:intfun}{A2}]
$\CLd:\CX\to\reals$ is $\lip$-Lipschitz for the $L^1$ metric on $\CX$:
\[
\bigl|\CLd(X)-\CLd(Y)\bigr|
\le
\lip\|X-Y\|_1
=
\lip\int_0^T\bigl|X(t)-Y(t)\bigr|\,\mathrm{d}t,
\qquad
\forall X,Y\in\CX.
\]

\item[\nl{as:uh}{A3}]
The Stratonovich fields of~\eqref{eq:strat_sde} satisfy a uniform
H\"ormander condition: the Lie algebra generated by
$V_0^\theta,\ldots,V_{\bdim}^\theta$ spans $\reals^{\xdim}$ uniformly in
$(x,\theta)$.
(A sufficient condition is uniform ellipticity of
$\sigma_{\theta}(\sigma_{\theta})^\top$.)
\end{myassumptions}

\ref{as:vecbd} is the standard unique-strong-solution setup~\cite{kloeden1992numerical}, with a uniform bound that is the working hypothesis of~\cite{litterer2012high} and holds on compact sets
that trap $X^\theta$ on $[0,T]$ with high probability. \ref{as:intfun} is the structural assumption on~\eqref{eq:objective}; it covers common latent-SDE losses~\cite{li2020scalable}.
\ref{as:uh} supplies the density smoothness used in the estimates behind Theorem~\ref{thm:cubap}.

\subsection{Cubature path construction and error}
\label{sec:cpc}

The paths $\om_z$ and trajectories $\phisoln$ in~\eqref{eq:mto} are built as follows.
Fix a local Lyons--Victoir rule of degree $m$ at time $1$, with $\cn$ paths $\{\om_j\}_{j\in[\cn]}$ and weights $\{\lambda_j\}_{j\in[\cn]}$ as in
Definition~\ref{def:cub} ($\cn=\wcn$ of  Section~\ref{sec:wsc_prelim}). Explicit nodes exist for $m=3,5$~\cite{lyons2004cubature} and $m=7$~\cite{ferrucci2026high}.
Partition
\begin{equation}
\label{eq:discint}
0=t_0<t_1<\cdots<t_k=T,
\qquad
s_i:=t_i-t_{i-1}.
\end{equation}
For $u\in[0,s]$ write the Brownian scaling
\[
\om_{s,j}^0(u)=u,
\qquad
\om_{s,j}^{\ell}(u)=\sqrt{s}\,\om_j^{\ell}(u/s),
\quad \ell=1,\ldots,\bdim.
\]
Index concatenated paths by
$I_z=(I_{z,1},\ldots,I_{z,k})\in[\cn]^k$.
The path $\om_z$ on $[0,T]$ is the continuous concatenation of the local
pieces $\om_{s_i,I_{z,i}}$ on $[t_{i-1},t_i]$,
\begin{equation}
\label{eq:pathcon}
\om_z\big|_{[t_{i-1},t_i]}
\;=\;
\om_{s_i,I_{z,i}}(\,\cdot\,-t_{i-1}),
\end{equation}
and the associated weight is
$\lambda_z':=\prod_{i=1}^{k}\lambda_{I_{z,i}}$.
Let $\phisol$ solve
\begin{equation}
\label{eq:ode}
\mathrm{d}\phisol
=
\sum_{i=0}^{\bdim} V_i^{\theta}(\phisol)\,\mathrm{d}\om_z^i(t),
\qquad
\phi_z^{\theta}(0)=\zeta_{\theta}(\vvar),
\end{equation}
and write $\phisoln:=(\phisol)_{t\in[0,T]}\in\CX$.
(The construction is conditional on the initial draw $\vvar$;
\ref{as:vecbd} makes the bound below uniform in that draw.)
The cubature estimator before recombination is
\begin{equation}
\label{eq:sum}
\Phi_{\CLd}^{\theta}
\;:=\;
\sum_{z\in[\cn^k]}
\lambda_z'\,\CLd(\phisoln),
\end{equation}
which is~\eqref{eq:mto} with $n=\cn^k$.

\begin{theorem}[Cubature path approximation error~\cite{snow2025efficient}]
\label{thm:cubap}
In the notation of this subsection, take the graded mesh
$t_i=T\bigl(1-(1-i/k)^\gamma\bigr)$ with $0<\gamma<m-1$.
Under \ref{as:vecbd}--\ref{as:uh},
\begin{equation}
\label{eq:cpae}
\bigl|\CE[\CLd(\Xt)]-\Phi_{\CLd}^{\theta}\bigr|
\le
3\,\lip\,T^{1/2}\,K\,C(m,\gamma)\,k^{-\gamma/2},
\end{equation}
where $K$ is independent of $T$ and $C(m,\gamma)$ depends only on $m$ and
$\gamma$.
\end{theorem}

\begin{proof}
We sketch the argument; the full proof is in~\cite{snow2025efficient}.
By \ref{as:intfun} the left-hand side of~\eqref{eq:cpae} is at most $\lip$
times the $1$-Wasserstein distance between $P(\theta)$ and the discrete path measure $\sum_z\lambda_z'\delta_{\phisoln}$. Cylinder functionals are dense for that metric, so it is enough to control finite-dimensional (time-grid) weak error. On each piece of~\eqref{eq:discint} this is the Lyons--Victoir remainder of
Proposition~\ref{prop:cub_ode}; concatenating on the graded mesh and using \ref{as:uh} to pass from smooth test functions to Lipschitz path
functionals produces~\eqref{eq:cpae}.
\end{proof}

Thus $\CE[\CLd(\Xt)]$ can be approximated to arbitrary accuracy in the partition size $k$ by solving the ODE ensemble~\eqref{eq:ode}, which is
the constructive content of~\eqref{eq:mto} before recombination. Available explicit rules fix $m\in\{3,5,7\}$, hence the admissible
$\gamma$; $m=5$ already yields the training rates discussed in~\cite{snow2025efficient}.\footnote{The constraint $0<\gamma<m-1$ is why
higher-degree constructions~\cite{ferrucci2026high} tighten~\eqref{eq:cpae}.
Degrees $3$ and $5$ of~\cite{lyons2004cubature} suffice for the complexity
claims in Meta Theorem~\ref{thm:meta}.}

\textit{Remark (Wasserstein duality).}
The bound~\eqref{eq:cpae} is inherited from a $W_1$ estimate between $P(\theta)$ and the cubature path measure. By Kantorovich--Rubinstein duality, the same estimate applies to every path functional satisfying \ref{as:intfun}, not only to a linear readout of $X_T$.

\textit{Exponential support, then recombination.}
The raw estimator~\eqref{eq:sum} uses $\cn^k$ ODEs.
\cite{snow2025efficient} applies the recombination procedure of~\cite{litterer2012high} to reduce that support to a polynomial number $n$ of paths while keeping a controlled bias. For an explicit admissible tuple $(m,\gamma,r)$ the resulting error is $\CO(n^{-1})$ in the number of surviving ODE solves, which is the rate stated in Meta Theorem~\ref{thm:meta}. The trainer in Corollary~\ref{cor:mcor} differentiates the recombined sum. We do not repeat the recombination algorithm here.

\end{document}